\pdfoutput=1
\documentclass[10pt,twocolumn,letterpaper]{article}

\usepackage{iccv}
\usepackage{times}
\usepackage{epsfig}
\usepackage{graphicx}
\usepackage{amsmath}
\usepackage{amssymb}
\usepackage{tikz}
\usetikzlibrary{spy}
\usepackage{multirow}
\usepackage{subcaption}
\usepackage{overpic}
\usepackage{marvosym}

\usepackage{booktabs}

\usepackage[accsupp]{axessibility} % Improves PDF readability for those with disabilities.

\usepackage[breaklinks=true,bookmarks=false]{hyperref}

\iccvfinalcopy % *** Uncomment this line for the final submission

\def\iccvPaperID{6077} % *** Enter the ICCV Paper ID here

\begin{document}

\title{Generalizing Event-Based Motion Deblurring in Real-World Scenarios}
% \title{Generalizing Event-based Motion Deblurring in the Real World}

\author{Xiang Zhang\textsuperscript{1},$\ $  Lei Yu\textsuperscript{1\Letter},$\ $  Wen Yang\textsuperscript{1},$\ $ Jianzhuang Liu\textsuperscript{2},$\ $ Gui-Song Xia\textsuperscript{1}
\\
\textsuperscript{1}Wuhan University
\quad
% \textsuperscript{2}Huawei Noah’s Ark Lab
\textsuperscript{2}Shenzhen Institute of Advanced Technology
\\
{\tt\small \{xiangz, ly.wd, yangwen, guisong.xia\}@whu.edu.cn, 
% liu.jianzhuang@huawei.com}
jz.liu@siat.ac.cn} 
}

% \and

% \and

\maketitle
% \thispagestyle{empty}

%%%%%%%%% ABSTRACT
\begin{abstract}
Event-based motion deblurring has shown promising results by exploiting low-latency events. However, current approaches are limited in their practical usage, as they assume the same spatial resolution of inputs and specific blurriness distributions. This work addresses these limitations and aims to generalize the performance of event-based deblurring in real-world scenarios. We propose a scale-aware network that allows flexible input spatial scales and enables learning from different temporal scales of motion blur. A two-stage self-supervised learning scheme is then developed to fit real-world data distribution. By utilizing the relativity of blurriness, our approach efficiently ensures the restored brightness and structure of latent images and further generalizes deblurring performance to handle varying spatial and temporal scales of motion blur in a self-distillation manner. Our method is extensively evaluated, demonstrating remarkable performance, and we also introduce a real-world dataset consisting of multi-scale blurry frames and events to facilitate research in event-based deblurring. 
% Codes and datasets are available at: \url{https://github.com/XiangZ-0/GEM}.

\section*{Multimedia Material}
\rm{The Multi-Scale Real-world Blurry Dataset (MS-RBD) and our Pytorch implementation are available at: \url{https://github.com/XiangZ-0/GEM}.}

\end{abstract}

%%%%%%%%% BODY TEXT
\section{Introduction}
\footnotetext{\textsuperscript{\Letter}Corresponding author}
\footnotetext{The research was partially supported by the National Natural Science Foundation of China under Grants 62271354 and 61871297. 
}
Due to the fixed exposure time of frame-based cameras, motion blur often occurs in scenes with dynamic targets or camera ego-motion, degrading the quality of the acquired images \cite{koh2021single,zhang2022deep}. Conventional motion deblurring approaches attempt to resolve this by exploiting deconvolution and blur kernel estimation techniques \cite{xu2010two,krishnan2011blind}, and recent research further improves the deblurring performance with the advanced deep-learning methods \cite{jin2018learning,zhang2021exposure}. However, traditional frame-based methods usually assume specific motion patterns, \eg, linear or quadratic motion trajectory, for blurry images and thus often face challenges in real-world scenarios with complex non-uniform motions. In addition, due to the motion ambiguity and texture erasure issues in blurry images \cite{red_xu2021motion,zhang2022unifying}, frame-based approaches often struggle to extract the precise motion and restore the accurate latent images from severely blurred frames.

\begin{figure}[t]
	\centering
	\includegraphics[width=\linewidth]{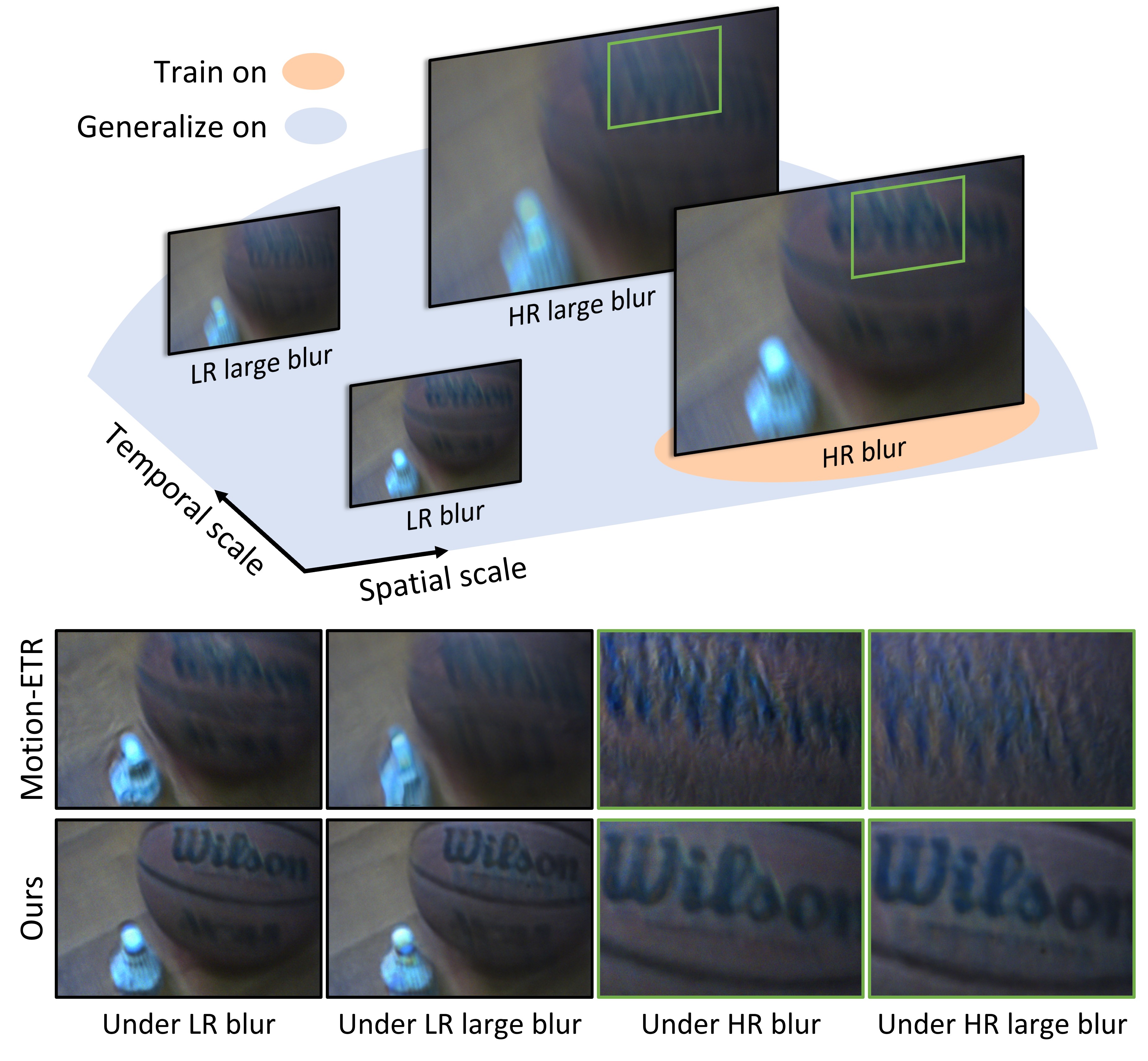}
        % \vspace{-1em}
	\caption{ An illustrative example of motion deblurring via the state-of-the-art algorithm Motion-ETR \cite{zhang2021exposure} and our proposed method, which is trained on HR blurry frames and LR events in a self-supervised manner and can generalize to the inputs at different temporal and spatial scales. }
    % \vspace{-1em}
	\label{fig:first}
\end{figure}

\par
% low latency for motion, brightness change for high-contrast texture
The advent of event cameras poses a paradigm shift in visual perception and information acquisition, benefiting a wide variety of applications \cite{lichtsteiner128Times1282008,gallego2020event,rebecq2019high,zhang2021event,hidalgo2022event,zhang2022formulating,ESAIPAMI,gao2022superfast,eslplusplus}. For motion deblurring tasks, the microsecond-level low latency of events enables almost continuous observation of dynamic scenes and alleviates the motion ambiguity in blurry frames \cite{edi_pan2019bringing,esl_wang2020event}. Moreover, the brightness changes recorded in event streams inherently correspond to high-contrast edges, compensating for the intensity texture erased by motion blur \cite{red_xu2021motion,zhang2022unifying,efnet,uevd}.
However, the performance of current event-based deblurring methods is usually confined to the distribution of training data, \eg, frames with a certain range of blurriness and the same spatial resolution as events, posing limitations in real-world scenarios.
% and thus poses limitations when applied in real-world scenarios. Specifically, 
\begin{itemize}
    % \vspace{-1mm}
    \item \textbf{Temporal Limitation:} Most previous approaches synthesize or collect blurry frames in a fixed range of exposure time for training \cite{esl_wang2020event,red_xu2021motion}, which implicitly assumes motion blur with a specific distribution of blurriness.
     However, real-world motion blur often violates this assumption in highly dynamic scenes, resulting in a performance drop of pre-trained models.
     % \vspace{-1mm}
    \item \textbf{Spatial Limitation:} Existing methods mainly
    take frames and events of the same spatial resolution as input, ignoring that frame-based cameras usually have larger spatial resolution than event-based ones in practice \cite{gallego2020event}. Besides, due to the varying distributions of events at different spatial scales \cite{gehrig2022high}, how to effectively deblur High-Resolution (HR) frames with Low-Resolution (LR) events remains an open problem.
    % \vspace{-1mm}
\end{itemize}

\par 
In this paper, we propose to address the above issues and generalize the performance of event-based motion deblurring in both spatial and temporal domains, as shown in Fig.~\ref{fig:first}.
% In detail, we first present a unified formulation of blurry and sharp latent images, and propose an exposure-guided event representation to enable 
In detail, a Scale-Aware Network (SAN) is first designed to extract high frame-rate HR sequences from a single HR blurry frame and its concurrent LR events. Inspired by implicit neural representation \cite{chen2021learning}, we implement a Multi-Scale Feature Fusion (MSFF) module to represent frame and event features in a spatially continuous manner, which allows flexible setups of input spatial resolutions. 
In the temporal dimension, an Exposure-Guided Event Representation (EGER) is presented to enable the arbitrary selection of target latent images without requiring model modification or re-training. 
To fit real-world data distribution, a two-stage self-supervised learning framework is further proposed. In the first stage, we efficiently supervise the restored brightness and structure of latent images by utilizing the relativity of blurriness. Following that, a self-distillation strategy is applied to generalize the deblurring performance to handle varying spatial and temporal scales of motion blur. 
% We additionally record a Multi-Scale Real-world Blurry Dataset (MS-RBD) composed of real-world HR blurry frames and LR events using a beam splitter setup. 
% Experiments on both synthetic and real-world datasets demonstrate the favorable performance of our approach. 
Overall, our contributions are three-fold:
\begin{itemize}
    % \vspace{-2mm}
    \item A scale-aware network is presented to allow flexible setups of input spatial resolutions and output temporal scales, which is able to restore high frame-rate HR sequences from HR blurry frames and LR events.
    % \vspace{-1.5mm}
    \item 
    A two-stage self-supervised learning framework is proposed to efficiently fit real-world data distributions and generalize deblurring performance to handle varying spatial and temporal scales of motion blur.
    % \vspace{-1.5mm}
    \item A real-world dataset MS-RBD containing HR blurry frames and LR events is built to facilitate deblurring research.  Extensive experiments on both synthetic and real datasets validate the effectiveness of our approach.
    % \vspace{-2mm}
\end{itemize}

% In detail, a Scale-Aware Network (SAN) is first designed to represent frames and events in a spatially continuous manner, which allows flexible setups of multi-scale inputs. 
% We also present an exposure-guided event representation to unify blurry and sharp latent images and enable arbitrary 
% latent image reconstruction without network re-training. 
% Furthermore, a self-supervised learning framework for event-based motion deblurring is proposed, where the deblurring performance is guaranteed by constraining brightness and structure consistency and generalized to different spatial resolutions of inputs and varying temporal scales of motion blur.
% Overall, our contributions are mainly three-fold:
% \begin{itemize}
%     \vspace{-2mm}
%     \item A scale-aware network is presented to allow flexible setups of input spatial resolutions and output temporal scales, which is able to restore high frame-rate video sequences from HR blurry frames and LR events.
%     \vspace{-2mm}
%     \item 
%     A self-supervised learning framework is proposed for efficient network training with real-world data and deblurring performance generalization to varying spatial and temporal scales of motion blur.
%     \vspace{-2mm}
%     \item A real-world dataset containing HR blurry frames and LR events is built to facilitate deblurring research.  Extensive experiments on both synthetic and real datasets validate the effectiveness of our proposed method.
%     \vspace{-2mm}
% \end{itemize}

\section{Related Work}
% \subsection{Motion Deblurring}
\noindent \textbf{Motion Deblurring.} How to recover sharp images from motion-blurred frames has been investigated for decades \cite{fergus2006removing,pan2016blind,xu2010two,krishnan2011blind,jin2018learning,zhang2021exposure,koh2021single,zhang2022deep}. Conventional deblurring methods often model the blurred image as a latent sharp image convolved with a blur kernel in the presence of additive noise \cite{fergus2006removing}, and several techniques have been adopted for motion deblurring, including deconvolution \cite{krishnan2011blind}, kernel estimation \cite{xu2010two}, and dark channel prior \cite{pan2016blind}. Recently, deep-learning approaches are also employed to achieve better deblurring results and extract video sequences from blurry frames \cite{jin2018learning,zhang2021exposure}. By exploiting an ordering-invariant constraint, LEVS gradually resumes the temporal ordering embedded in motion blur and recovers sharp sequences from a blurry input \cite{jin2018learning}. Motion-ETR further improves deblurring performance by utilizing 
Deformable Convolutional Networks (DCNs) \cite{zhu2019deformable} to predict the motion trajectory within blurry frames, which tackles temporal disorder and enables the recovery of non-linear exposure trajectories \cite{zhang2021exposure}.

\par 
However, traditional frame-based methods usually assume specific motion patterns of blurry frames and thus often fail in real-world scenarios with complex non-uniform motion. Besides, large motion blur will eliminate the intensity texture in the acquired frames, posing challenges to recovering satisfied latent images from blurry inputs. 

% \subsection{Event-based Motion Deblurring}
\noindent \textbf{Event-based Motion Deblurring.}
Recent works have revealed the advantages of events in motion deblurring \cite{edi_pan2019bringing,esl_wang2020event,red_xu2021motion,song2022cir,zhang2022unifying,efnet,uevd,eslplusplus}. With the low latency and high temporal resolution of event cameras \cite{gallego2020event}, events naturally encode the information of high-contrast texture and precise motion of dynamic scenes, facilitating the reconstruction of sharp latent images under complex motion. Previous work of \cite{edi_pan2019bringing} first establishes the Event-based Double Integral (EDI) model for motion deblurring, which bridges the blurry frames and latent sharp images with events. Following that, learning-based methods are developed to achieve better results by adopting techniques like sparse coding \cite{esl_wang2020event,eslplusplus}, parametric polynomial \cite{song2022cir}, and cross-modal attention \cite{efnet}. To fit real-world data distribution, recent works also focus on learning from real blurry frames and events by semi-/self-supervised methods \cite{red_xu2021motion, zhang2022unifying}.

\par
Although event-based methods have made significant progress in motion deblurring, the aforementioned approaches generally focus on deblurring frames with specific temporal scales of motion blur and the same spatial resolutions as events, showing limitations in real-world applications.
% Even though event-based methods have made significant progress in motion deblurring, the aforementioned approaches still have limitations in both temporal and spatial domains, where the deblurring performance is confined to a specific distribution of blur related to training data, and the resolution of frames assumed to be the same as that of events.
In our approach, a scale-aware network is designed to deblur HR frames with LR events and simultaneously enable flexible setups of input spatial resolutions. Moreover, a self-supervised learning framework is proposed to efficiently fit real-world data distribution and generalize the deblurring performance in both spatial and temporal domains.

\section{Method}
In this section, we first formulate event-based deblurring and our goal in Sec.~\ref{sec:probform}. Based on this, we then introduce the scale-aware network in Sec.~\ref{sec:san} and finally propose our self-supervised learning method in Sec.~\ref{sec:ssl}.

% We first formulate event-based deblurring and our goal in Sec.~\ref{sec:probform}. Based on the target, we then introduce the scale-aware network in Sec.~\ref{sec:san}  to enable flexible input spatial resolutions and allow learning from different temporal scales of motion blur. In Sec.~\ref{sec:ssl}, a two-stage self-supervised learning method is finally proposed to train our model with real-world data and generalize its deblurring performance in both temporal and spatial dimensions.

\subsection{Problem Formulation}\label{sec:probform}
We first review the basic model of event-based motion deblurring, which aims to restore sharp latent images from blurry frames and the corresponding events. 
% The task of event-based motion deblurring is to restore latent sharp images from blurry frames and the corresponding events. 
According to the event generation model \cite{lichtsteiner128Times1282008,gallego2020event}, each event is emitted asynchronously whenever the log-scale brightness change reaches the event threshold $c>0$, \begin{equation}\label{eq:event_model}
    \operatorname{log}(I(t,\mathbf{x})) - \operatorname{log}(I(f,\mathbf{x})) = p\cdot c,
\end{equation}
where $\operatorname{log}(I(t,\mathbf{x})), \operatorname{log}(I(f,\mathbf{x}))$ correspond to the log-scale intensity of pixel $\mathbf{x}$ at time $t$ and $f$, and $p\in\{+1,-1\}$ denotes the polarity showing the direction of brightness change. On the other hand, blurry frames can be formulated as the average of the latent images within the exposure period $\mathcal{T}$ \cite{chen2018reblur2deblur} (pixel position $\mathbf{x}$ is omitted for readability),
\begin{equation}\label{eq:blur_latent}
    B_T = \frac{1}{T} \int_{t\in \mathcal{T}} I(t) dt,
\end{equation}
where $B_T$ indicates the blurry frame captured with exposure time $T$. Combining Eq.~\eqref{eq:event_model} and \eqref{eq:blur_latent}, one can bridge blurry frames and sharp images by the EDI model \cite{edi_pan2019bringing},
\begin{align}  
    I(t) &= \frac{B_T}{E(t,\mathcal{T})}
    % B_T &= I(t)\cdot E(t,\mathcal{T})
    ,\ \text{with} \label{eq:edi}
    \\
    E(t,\mathcal{T}) &= \frac{1}{T} \int_{f \in \mathcal{T}} \operatorname{exp} (c\int_t^f e(s)ds)df, \label{eq:edi_integral}
\end{align}
where $e(\tau)\triangleq p \cdot \delta(\tau-t)$ indicates the continuous event representation and $\delta(\cdot)$ denotes the Dirac function. 
Since directly restoring $I(t)$ via Eq.~\eqref{eq:edi} often suffers from the instability of event threshold $c$ in practice \cite{gallego2020event,red_xu2021motion}, learning-based approaches are employed to better fit the statistics of events \cite{esl_wang2020event, red_xu2021motion}, which are generally in the form of 
% Although one can directly restore the sharp latent image via Eq.~\eqref{eq:edi}, the deblurring performance is often limited due to the spatial and temporal instability of event threshold $c$ in practice \cite{gallego2020event,red_xu2021motion}. Thus, learning-based approaches are employed to better fit the statistics of events \cite{esl_wang2020event, red_xu2021motion}, and they are generally in the form of 
\begin{equation}\label{eq:EMD}
    I(t) = \operatorname{Deblur}(t;B_T,\mathcal{E}_\mathcal{T}),\quad \forall t\in \mathcal{T},
\end{equation}
where $\operatorname{Deblur}(\cdot)$ denotes a motion deblurring network and $\mathcal{E}_\mathcal{T}$ indicates the events triggered within $\mathcal{T}$. 

\par 
Define the spatial resolution ratio of frames to events as  $\mathcal{R}(B_T,\mathcal{E}_\mathcal{T})$, \eg, $\mathcal{R}(B_T,\mathcal{E}_\mathcal{T})=4$ means the resolution of frame $B_T$ is four times that of events $\mathcal{E}_\mathcal{T}$, previous learning-based approaches are commonly trained on the dataset
\begin{equation}\label{eq:train_data}
\mathcal{D}(\mathbf{T}, \mathbf{R}) \triangleq \{B_T, \mathcal{E}_\mathcal{T} | T \in  \mathbf{T},\ \mathcal{R}(B_T,\mathcal{E}_\mathcal{T})\in \mathbf{R}\}
\end{equation}
with $\mathbf{R}=\{1\}$ indicating the same spatial resolution of frames and events, and $\mathbf{T}=\{T_k\}_{k=1}^K$ denoting a set composed of $K$ exposure parameters. 
Once trained, it is difficult to directly apply previous methods to process real-world inputs with $\mathcal{R}(B_T,\mathcal{E}_\mathcal{T})>1$, \ie, HR blurry frames and LR events. Besides, the set $\mathbf{T}$ implicitly assumes a specific distribution of blurriness, which often results in a performance drop of pre-trained models when inferring more blurred frames. 
% Hence, the applicable range of existing approaches is limited by $\mathbf{T}$ and $\mathbf{R}$.

\par 
To foster the application of event-based motion deblurring in real-world scenarios, it is necessary to enlarge the sets of $\mathbf{T}$ and $\mathbf{R}$. However, collecting sufficient datasets to cover a wide range of $\mathbf{T}$, $\mathbf{R}$ is time-consuming and impractical. Also, sharp ground-truth images are difficult to collect when recording real-world blurry datasets and thus are usually unavailable for training. Therefore, the goal of our work is to design a fully self-supervised deblurring algorithm that only needs to train on a dataset $\mathcal{D}(\mathbf{T}, \{\bar{R}\})$ with $\forall \bar{R}\geq1$ to fit real-world setups, but is able to generalize on a larger set $\mathcal{D}(\mathbf{T}^{*}(M), \mathbf{R}^{*}(\bar{R}))$ as shown in Fig.~\ref{fig:first}, where
\begin{equation}\label{eq:real_data}
\begin{aligned}
    &\mathbf{T}^{*}(M) \triangleq \sum_{m=1}^M \{mT_k\}_{k=1}^K,
    \\
    &\mathbf{R}^{*}(\bar{R}) \triangleq \{R| 1 \leq R \leq \bar{R},R\in \mathbb{R} \},
\end{aligned}
\end{equation}
with $M\in\mathbb{N}^+$ denoting a parameter that can be chosen to determine the temporal scale of motion blur.

\begin{figure*}[t]
	\centering
	\begin{subfigure}{0.214\linewidth}
	\centering
		\includegraphics[width=\linewidth]{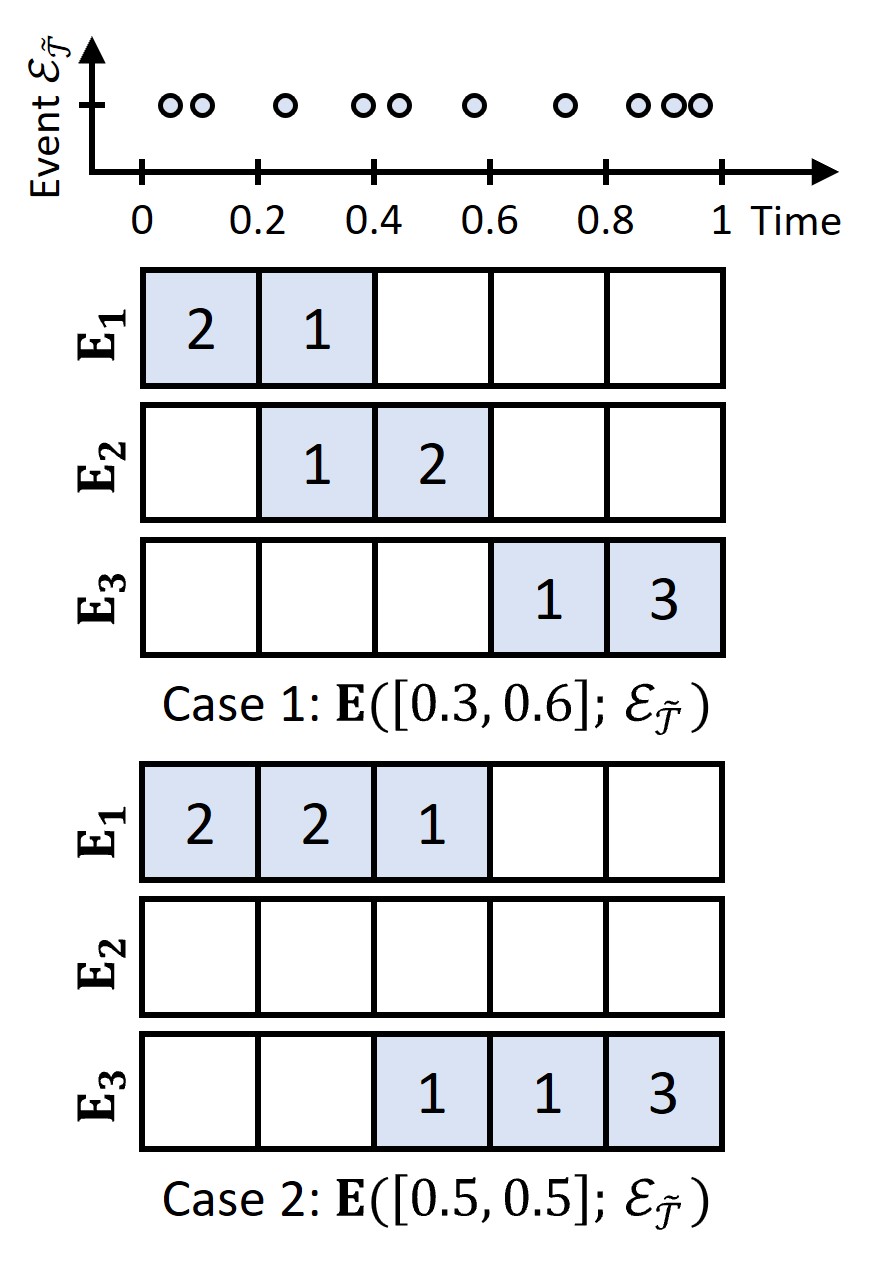}
		% \vspace{-1.5em}
		\caption{\rmfamily \fontsize{8pt}{0} Event representation}
		\label{fig:network-eger}
	\end{subfigure}
	\begin{subfigure}{0.73\linewidth}
	\centering
		\includegraphics[width=\linewidth]{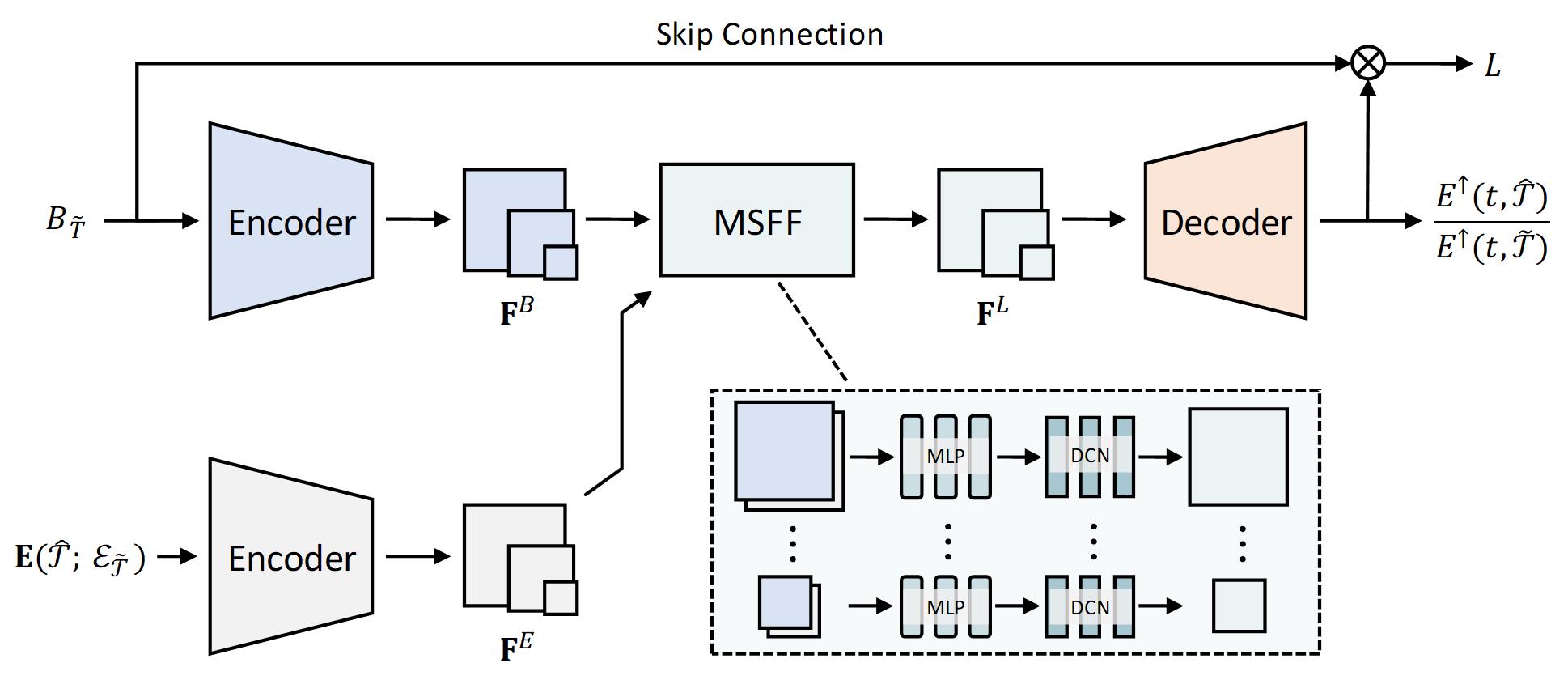}
		\caption{\rmfamily \fontsize{8pt}{0} Scale-aware network}
		\label{fig:network-arch}
	\end{subfigure}
        % \vspace{-0.5em}
	\caption{
% 	(a) Exposure-guided event representation and (b) scale-aware network. 
	(a) An example of our Exposure-Guided Event Representation (EGER). The event stream $\mathcal{E}_{\tilde{\mathcal{T}}}$ contains 10 negative events in $\tilde{\mathcal{T}}=[0,1]$. We show two cases of $\mathbf{E}(\hat{\mathcal{T}};\mathcal{E}_{\tilde{\mathcal{T}}})$ under $N=5$ with $\hat{\mathcal{T}}=[0.3,0.6]$ and $\hat{\mathcal{T}}=[0.5,0.5]$, and the operation is the same for positive events. (b) Structure of our proposed network with the Multi-Scale Feature Fusion (MSFF) module.
 % \textcolor{red}{Two encoders are first employed to extract multi-scale features $\mathbf{F}^B,\mathbf{F}^E$ from HR blurry frames $B_{{T}}$ and LR events $\mathbf{E}(\hat{\mathcal{T}};\mathcal{E}_{\mathcal{T}})$. Then the multi-scale features $\mathbf{F}^B,\mathbf{F}^E$ are represented in a spatially continuous manner and fused by our MSFF module. Finally, the latent image $L$ is generated by decoding the fused feature $\mathbf{F}^L$.} 
 }
	\label{fig:network}
    % \vspace{-1em}
\end{figure*}
\subsection{Scale-Aware Network}\label{sec:san}
Unlike previous methods that focus on fitting Eq.~\eqref{eq:edi}, our Scale-Aware Network (SAN) aims to approximate a more general function to allow flexible input spatial scales and enable learning from different temporal scales of motion blur.
% Unlike previous methods that fit the function of Eq.~\eqref{eq:edi}, our Scale-Aware Network (SAN) allows flexible setups of input spatial scales and enables learning from different temporal scales of motion blur.
% we propose to approximate a more general case and learn motion deblurring by exploiting the relativity of blurriness.
% via our Scale-Aware Network (SAN).
% The design of our Scale-Aware Network (SAN) is based on a generalized version of the EDI model, \ie, Eq.~\eqref{eq:edi}.
Due to the different spatial scales of frames and events in our task, we first modify Eq.~\eqref{eq:edi} to 
% In our task with different spatial scales of frames and events, we first modify Eq.~\eqref{eq:edi} to 
\begin{equation}\label{eq:modified_edi}
I(t) = \frac{B_T}{E^{\uparrow}(t,\mathcal{T})},
\end{equation}
with $E^{\uparrow}(t,\mathcal{T})$ indicating the upsampled version of $E(t,\mathcal{T})$ to match the spatial resolution of $B_T$.
Next we consider a more blurred frame $B_{\tilde{T}}$ and similarly get $I(t) = {B_{\tilde{T}}}/{E^{\uparrow}(t,\tilde{\mathcal{T}})}$ with $T<\tilde{T}$ and $\mathcal{T}\subset\tilde{\mathcal{T}}$. 
For the same target image $I(t)$, one can then derive 
% a general version of EDI model \cite{edi_pan2019bringing},
% a general function of Eq.~\eqref{eq:modified_edi},
% the following relation:
\begin{equation}\label{eq:uni_form}
    B_T = \frac{E^{\uparrow}(t,\mathcal{T})}{E^{\uparrow}(t,\tilde{\mathcal{T}})} B_{\tilde{T}},
\end{equation}
% which is a general formulation that unifies sharp and blurry latent images. 
which converts the more blurred frame $B_{\tilde{T}}$ into its less blurred latent image $B_{{T}}$. 
Inspired by this, we design our SAN to approximate a general function
% Our SAN is designed to approximate Eq.~\eqref{eq:uni_form} to take into account different input spatial scales and enable learning from different temporal scales of motion blur,
\begin{equation}\label{eq:san}
    L = \frac{E^{\uparrow}(t,\hat{\mathcal{T}})}{E^{\uparrow}(t,\tilde{\mathcal{T}})} B_{\tilde{T}} \approx \operatorname{SAN}(\hat{\mathcal{T}}; B_{\tilde{T}}, \mathcal{E}_{\tilde{\mathcal{T}}}),
\end{equation}
where $\hat{\mathcal{T}}\subset \tilde{\mathcal{T}}$ controls the output temporal scale of the target latent image $L$. Thus, our SAN is able to restore both sharp and blurry latent images by setting different $\hat{\mathcal{T}}$, \ie,
% by choosing different $\hat{\mathcal{T}}$, our SAN is capable of
\begin{itemize}
    % \vspace{-1mm}
    \item \textbf{Blur2sharp conversion:} If $\hat{\mathcal{T}}=[t,t]$, $E^{\uparrow}(t,\hat{\mathcal{T}})=1$ holds since no event is integrated, and thus $L=I(t)$.
    % \vspace{-1mm}
    \item \textbf{Blur2blur conversion:} If $\hat{\mathcal{T}}=\mathcal{T}$, the target function in Eq.~\eqref{eq:san} becomes Eq.~\eqref{eq:uni_form}, and thus $L=B_T$.
    % \vspace{-1mm}
\end{itemize}
This enables SAN to learn from blur2blur conversion without requiring sharp ground-truth images. Moreover, our SAN does not assume the same spatial resolution of inputs. To fulfill the temporal and spatial flexibility, an Exposure-Guided Event Representation (EGER) and a Multi-Scale Feature Fusion (MSFF) module are respectively proposed.

\par 
\noindent \textbf{Exposure-Guided Event Representation.} 
The goal of EGER is to explicitly model the conversion relationship between the input blurry frame and the latent image with events, which can be regarded as preparing events for computing $E^{\uparrow}(t,\hat{\mathcal{T}})/E^{\uparrow}(t,\tilde{\mathcal{T}})$ in Eq.~\eqref{eq:san}.
Given an event stream $\mathcal{E}_{\tilde{\mathcal{T}}}$ with $\tilde{\mathcal{T}}\triangleq[{t}_s,{t}_e]$ and the target exposure period $\hat{\mathcal{T}}\triangleq[\hat{t}_s,\hat{t}_e]$, we first evenly divide $\tilde{\mathcal{T}}$ into $N$ temporal bins and generate three $2N\times H \times W$ event tensors $\mathbf{E}_1, \mathbf{E}_2,$ and $\mathbf{E}_3$ with $2, H, W$ indicating event polarity, height, and width. 
% Denoted by $\mathbf{E}_1, \mathbf{E}_2,$ and $\mathbf{E}_3$, 
The three tensors $\mathbf{E}_1, \mathbf{E}_2,$ and $\mathbf{E}_3$ accumulate the events split based on the intervals $[{t}_s, \hat{t}_s], [\hat{t}_s, \hat{t}_e],$ and $[\hat{t}_e, {t}_e]$, respectively. By simple event splitting, $\mathbf{E}_2$ contains events $\mathcal{E}_{\hat{\mathcal{T}}}$ in the target exposure period for computing $E^{\uparrow}(t,\hat{\mathcal{T}})$, and the combination of $\mathbf{E}_1, \mathbf{E}_2,$ and $\mathbf{E}_3$ corresponds to events $\mathcal{E}_{\tilde{\mathcal{T}}}$ for $E^{\uparrow}(t,\tilde{\mathcal{T}})$.
Then our EGER is formed by concatenating the three event tensors, \ie,
\begin{equation}\label{eq:eger}
    \mathbf{E}(\hat{\mathcal{T}};\mathcal{E}_{\tilde{\mathcal{T}}}) = \operatorname{Concat}(\mathbf{E}_1, \mathbf{E}_2, \mathbf{E}_3),
\end{equation}
where $\mathbf{E}(\hat{\mathcal{T}};\mathcal{E}_{\tilde{\mathcal{T}}})$ is the EGER of target exposure period $\hat{\mathcal{T}}$ conditioned on the input events $\mathcal{E}_{\tilde{\mathcal{T}}}$.

\par 
As the toy example shown in Fig.~\ref{fig:network-eger}, the input event stream can be represented as different $\mathbf{E}(\hat{\mathcal{T}};\mathcal{E}_{\tilde{\mathcal{T}}})$ according to the chosen $\hat{\mathcal{T}}$. 
This allows SAN to determine the output temporal scales and recover both blurry (\eg, case 1 in Fig.~\ref{fig:network-eger}) and sharp (\eg, case 2 in Fig.~\ref{fig:network-eger}) latent images from the same input. Also, EGER enables flexible selection of $\hat{\mathcal{T}}\subset{\tilde{\mathcal{T}}} $  for arbitrarily high frame-rate video generation.
% and also enables flexible selection of the target exposure time $\mathcal{T}\subset\tilde{\mathcal{T}} $ for arbitrarily high-frame-rate video generation.

% This allows our SAN to recover both blurry (\eg, case 1 in Fig.~\ref{fig:network-eger}) and sharp (\eg, case 2 in Fig.~\ref{fig:network-eger}) latent images from the same input data, and also enables flexible selection of the target exposure time $\mathcal{T}\subset\tilde{\mathcal{T}} $ for arbitrarily high-frame-rate video generation.

% Denoted by $\mathbf{E}_1, \mathbf{E}_2,$ and $\mathbf{E}_3$, the three event tensors split the input events $\mathcal{E}_{\tilde{\mathcal{T}}}$ into three segments and accumulate the events based on intervals $[\tilde{t}_s, t_s], [t_s, t_e],$ and $[t_e, \tilde{t}_e]$, respectively.

\par 
\noindent \textbf{Multi-Scale Feature Fusion.}
Another challenge for SAN is the different spatial resolutions between HR blurry frames and LR events.
Inspired by the Local Implicit Image Function (LIIF) \cite{chen2021learning} that represents images in a spatially continuous manner, we propose to fuse frames and events by learning a continuous feature representation. As depicted in Fig.~\ref{fig:network-arch}, we first extract multi-scale blur and event features $\mathbf{F}^{B}=\{F^{B}_i\}, \mathbf{F}^{E}=\{F^{E}_i\}$ with $F^{B}_i, F^{E}_i$ denoting the features at the $i$-th scale by two encoder networks (our encoder and decoder networks are split from an hourglass network, detailed in the supplementary material). 
Considering the cross-sensor gap between frame-based and event-based cameras, we use the blur features to provide brightness reference and guide the upsampling of event features in our MSFF module. Specifically, a Multi-Layer Perceptron (MLP) is employed to predict the fused feature value from cross-modal local features, \ie,
$
    f_i(z) = \operatorname{MLP}_i(z, s ;F^{B}_i, F^{E}_i),
$
where $z$ indicates a 2D coordinate in the continuous spatial domain, $f_i(z)$ is the predicted feature at $z$, and $s=[s_h,s_w]$ is the size of the target feature pixel.
% Then $\mathbf{F}^{B}$ and $\mathbf{F}^{E}$ are fed into the MSFF module composed of a Multi-Layer Perceptron (MLP) and a DCN \cite{zhu2019deformable} at each scale. 
% Considering the cross-sensor gap between frame-based and event-based cameras, we feed both image and event features $F^{B}_i, F^{E}_i$ into the MLP to facilitate feature prediction, \ie, $
%     f_i(z) = \operatorname{MLP}_i(z, s ;F^{B}_i, F^{E}_i),
% $
% % \end{equation}
% where $z$ indicates a 2D coordinate in the continuous spatial domain, $f_i(z)$ is the predicted feature at $z$, and $s=[s_h,s_w]$ is the size of the target feature pixel.
% -----------------
% The MLP receives the target coordinates and the related local features as input, and predicts the feature value at target coordinates, \ie,
% % \begin{equation}\label{eq:mlp}
% $
%     f_i(z) = \operatorname{MLP}_i(z, s ;F^{B}_i, F^{E}_i),
% $
% % \end{equation}
% where $z$ indicates a 2D coordinate in the continuous spatial domain, $f_i(z)$ is the predicted feature at $z$, and $s=[s_h,s_w]$ is the size of the target feature pixel. 
% Considering the cross-sensor gap between frame-based and event-based cameras, both image and event features $F^{B}_i, F^{E}_i$ are fed into the MLP to facilitate feature prediction. 
Afterward, the coarsely fused feature is refined by a DCN with a larger receptive field, generating the final feature of latent image $F^{L}_i=\operatorname{DCN}_i(f_i)$. 
% Finally, we pass the features $\mathbf{F}^{L} = \{F^{L}_i\}$ to a decoder, and generate the latent image $L$ 
We finally pass the features $\mathbf{F}^{L}$ through a decoder network and restore the latent image $L$.

% Finally, we restore the latent image $L$ from the input blurry frame $B_T$ and the output of a decoder fed with features $\mathbf{F}^{L} = \{F^{L}_i\}$.

% Finally, a decoder is employed to restore the latent image $L$ from the features $\mathbf{F}^{L} = \{F^{L}_i\}$.
\par 
By utilizing the MSFF module, our SAN is able to effectively fuse the information of frames and events at different spatial resolutions. In addition, since the coordinates are continuous, MSFF enables flexible setups of input spatial scales, \eg, our SAN can simultaneously take inputs of $\mathcal{R}(B_{{T}}, \mathcal{E}_{{\mathcal{T}}})=4$ and $\mathcal{R}(B_{{T}}, \mathcal{E}_{{\mathcal{T}}})=2.5$ without network modification or re-training, facilitating practical usage.

\subsection{Self-Supervised Learning}\label{sec:ssl}
Our self-supervised learning approach consists of two stages: 
we first constrain the restored brightness and structure of latent images by utilizing the relativity of blurriness, and then generalize the deblurring performance in both temporal and spatial dimensions via self-distillation techniques.

% By utilizing the relativity of blurriness, we propose a two-stage self-supervised learning framework, where the deblurring performance is first ensured by constraining brightness and structure consistency, and then generalized to handle varying temporal and spatial scales of motion blur.

\noindent \textbf{Brightness and Structure Consistency.} 
Based on Eq.~\eqref{eq:san}, we propose to constrain the reconstruction brightness by learning blur2blur conversion. 
% Given a blurry video $\textbf{B}_T=\{B_{T}^{i}\}_{i=1}^{N_B}$ with $N_B$ blurry frames, we synthesize the more blurred sequence $\textbf{B}_{\tilde{T}}=\{B_{\tilde{T}}^{j}\}_{j=1}^{N_B/M}$ by averaging $M$ consecutive blurry frames ($M=2$ in our work), \ie, $B_{\tilde{T}}^{j}=\frac{1}{M}\sum_{i=M(j-1)+1}^{Mj}B_{T}^{i}$. 
Given $B_T$ from a blurry video, we synthesize a more blurred image $B_{\tilde{T}}$ by averaging $M$ adjacent blurry frames of $B_T$ ($M=2$ in our experiments) and formulate the constraint as
\begin{equation}\label{eq:loss_bc}
    \mathcal{L}_{BC} = \|B_T - \operatorname{SAN}(\mathcal{T};B_{\tilde{T}}, \mathcal{E}_{\tilde{\mathcal{T}}}) \|_1,
\end{equation}
which efficiently ensures brightness consistency by learning to restore $B_T$ from $B_{\tilde{T}}$.
\par 
According to Eq.~\eqref{eq:modified_edi}, recovering the structure of sharp latent images is equivalent to estimating accurate $E^{\uparrow}(t,\mathcal{T})$ for each $I(t)$. To achieve this, we first breakdown our SAN into $\operatorname{SAN}(\hat{\mathcal{T}};B_{\tilde{T}}, \mathcal{E}_{\tilde{\mathcal{T}}}) = \operatorname{SAN}^{E}(\hat{\mathcal{T}};B_{\tilde{T}}, \mathcal{E}_{\tilde{\mathcal{T}}})\cdot B_{\tilde{T}}$, where $\operatorname{SAN}^{E}(\cdot)$ estimates the event ratio based on Eq.~\eqref{eq:san} and Fig.~\ref{fig:network-arch},
\begin{equation}\label{eq:san_e}
   \frac{E^{\uparrow}(t,\hat{\mathcal{T}})}{E^{\uparrow}(t,\tilde{\mathcal{T}})} \approx \operatorname{SAN}^{E}(\hat{\mathcal{T}};B_{\tilde{T}}, \mathcal{E}_{\tilde{\mathcal{T}}}).
\end{equation}
By setting $\hat{\mathcal{T}}=[t,t]$, 
$\operatorname{SAN}^{E}(\cdot)$ is able to estimate $E^{\uparrow}(t,\tilde{\mathcal{T}})$ for restoring sharp latent image $I(t)$, \ie, 
% $\operatorname{SAN}^{E}(\cdot)$ is able to estimate $E^{\uparrow}(t,\mathcal{T})$ for restoring sharp image $I(t)$ by setting $\hat{\mathcal{T}}=[t,t]$, \ie,
\begin{equation}\label{eq:san_e_sharp}
    \frac{1}{E^{\uparrow}(t,\tilde{\mathcal{T}})} \approx \operatorname{SAN}^{E}([t,t];B_{\tilde{T}}, \mathcal{E}_{\tilde{\mathcal{T}}}).
\end{equation}
Then, we constrain the structure of the restored $I(t)$ by
\begin{equation}\label{eq:loss_sc}
    \mathcal{L}_{SC} = \left\| \operatorname{SAN}^{E}(\mathcal{T};B_{\tilde{T}}, \mathcal{E}_{\tilde{\mathcal{T}}})
    - \frac{\operatorname{SAN}^{E}([t,t];B_{\tilde{T}}, \mathcal{E}_{\tilde{\mathcal{T}}})}{\operatorname{SAN}^{E}([t,t];B_{{T}}, \mathcal{E}_{{\mathcal{T}}})} \right\|_1,
\end{equation}
where $\operatorname{SAN}^{E}(\mathcal{T};B_{\tilde{T}}, \mathcal{E}_{\tilde{\mathcal{T}}})$ provides strong supervision to avoid collapsing solutions as it is constrained in $\mathcal{L}_{BC}$. 
$\mathcal{L}_{SC}$ guarantees structure recovery by transferring the knowledge learned from blur2blur to the blur2sharp case. With $\mathcal{L}_{BC}$ and $\mathcal{L}_{SC}$, SAN efficiently achieves motion deblurring by ensuring the brightness and structure of sharp latent images.
% By exploiting $\mathcal{L}_{BC}$ and $\mathcal{L}_{SC}$, our SAN efficiently achieves motion deblurring by ensuring the brightness and structure of latent images.

% learns motion deblurring from only blurry pairs $\{B_T, B_{\tilde{T}}\}$.

% By learning to restore blurry latent images $B_T$ from a more blurred frame $B_{\bar{T}}$, $\mathcal{L}_{BC}$ efficiently ensures brightness consistency without supervision from ground-truth images. 

\begin{figure*}[t]
    \centering
    %% in 1
    \begin{subfigure}[b]{0.497\linewidth}
        \centering
        \begin{subfigure}[b]{0.511\linewidth}
        \centering
        \includegraphics[width=\linewidth, trim={8cm 0 4cm 0cm}, clip]{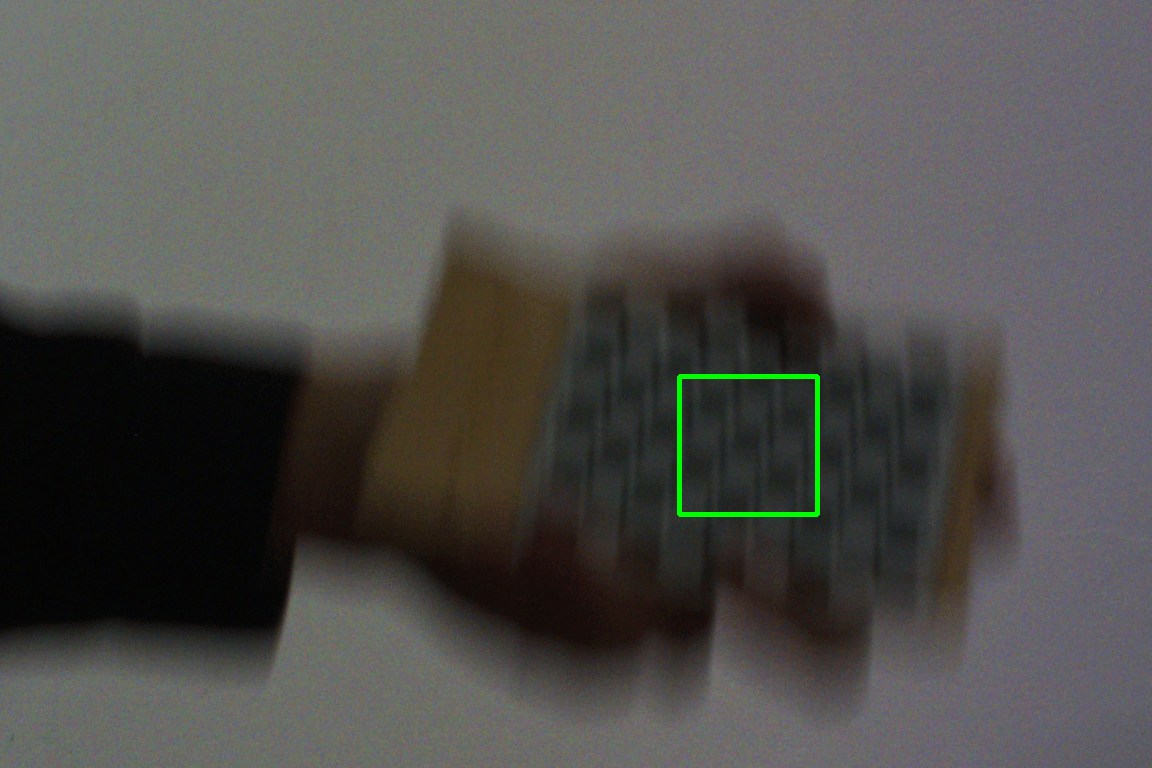}
        \vspace{-1.7em}
        \subcaption*{\centering Blurry image}
        \end{subfigure}
        \begin{subfigure}[b]{0.2195\linewidth}
        \centering
        \includegraphics[width=\linewidth, trim={0cm 0 0cm 0}]{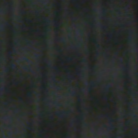}
        \vspace{-1.7em}
        \subcaption*{\centering LEVS}
        \vspace{-0.2em}
        \includegraphics[width=\linewidth, trim={0cm 0 0cm 0}]{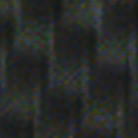}
        \vspace{-1.7em}
        \subcaption*{\centering EVDI+DASR}
        \end{subfigure}
        \begin{subfigure}[b]{0.2195\linewidth}
        \centering
        \includegraphics[width=\linewidth, trim={0cm 0 0cm 0}]{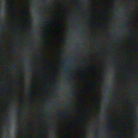}
        \vspace{-1.7em}
        \subcaption*{\centering Motion-ETR}
        \vspace{-0.2em}
        \includegraphics[width=\linewidth, trim={0cm 0 0cm 0}]{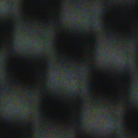}
        \vspace{-1.7em}
        \subcaption*{\centering Ours}
        \end{subfigure}
    \end{subfigure}
    %% in 2
    \begin{subfigure}[b]{0.497\linewidth}
        \centering
        \begin{subfigure}[b]{0.511\linewidth}
        \centering
        \includegraphics[width=\linewidth, trim={6cm 0 6cm 0cm}, clip]{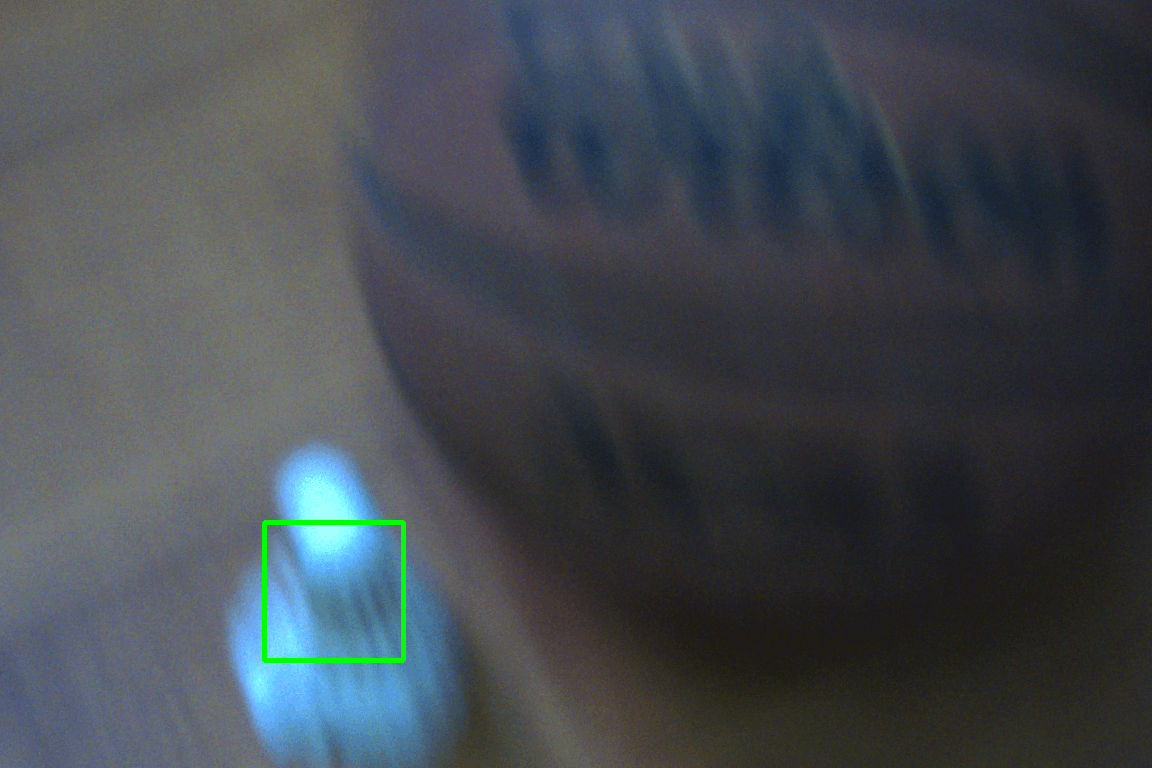}
        \vspace{-1.7em}
        \subcaption*{\centering Blurry image}
        \end{subfigure}
        \begin{subfigure}[b]{0.2215\linewidth}
        \centering
        \includegraphics[width=\linewidth, trim={0cm 0 0cm 0}]{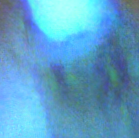}
        \vspace{-1.7em}
        \subcaption*{\centering LEVS}
        \vspace{-0.2em}
        \includegraphics[width=\linewidth, trim={0cm 0 0cm 0}]{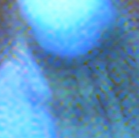}
        \vspace{-1.7em}
        \subcaption*{\centering EVDI+DASR}
        \end{subfigure}
        \begin{subfigure}[b]{0.2215\linewidth}
        \centering
        \includegraphics[width=\linewidth, trim={0cm 0 0cm 0}]{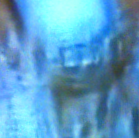}
        \vspace{-1.7em}
        \subcaption*{\centering Motion-ETR}
        \vspace{-0.2em}
        \includegraphics[width=\linewidth, trim={0cm 0 0cm 0}]{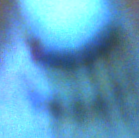}
        \vspace{-1.7em}
        \subcaption*{\centering Ours}
        \end{subfigure}
    \end{subfigure}
    \\
    %% out 1
    \begin{subfigure}[b]{0.497\linewidth}
        \centering
        \begin{subfigure}[b]{0.511\linewidth}
        \centering
        \includegraphics[width=\linewidth, trim={8cm 0 4cm 0cm}, clip]{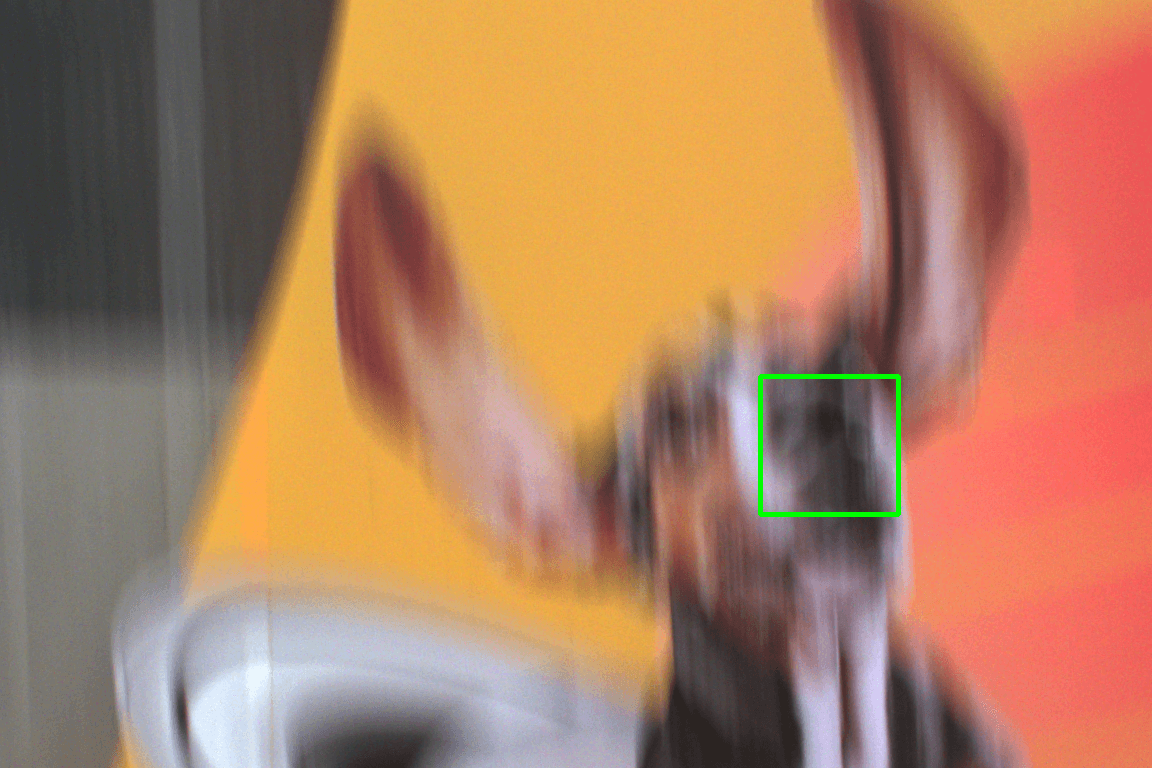}
        \vspace{-1.7em}
        \subcaption*{\centering Blurry image}
        \end{subfigure}
        \begin{subfigure}[b]{0.2195\linewidth}
        \centering
        \includegraphics[width=\linewidth, trim={0cm 0 0cm 0}]{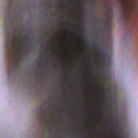}
        \vspace{-1.7em}
        \subcaption*{\centering LEVS}
        \vspace{-0.2em}
        \includegraphics[width=\linewidth, trim={0cm 0 0cm 0}]{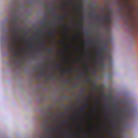}
        \vspace{-1.7em}
        \subcaption*{\centering EVDI+DASR}
        \end{subfigure}
        \begin{subfigure}[b]{0.2195\linewidth}
        \centering
        \includegraphics[width=\linewidth, trim={0cm 0 0cm 0}]{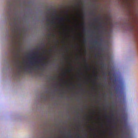}
        \vspace{-1.7em}
        \subcaption*{\centering Motion-ETR}
        \vspace{-0.2em}
        \includegraphics[width=\linewidth, trim={0cm 0 0cm 0}]{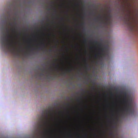}
        \vspace{-1.7em}
        \subcaption*{\centering Ours}
        \end{subfigure}
    \end{subfigure}
    %% out 2
    \begin{subfigure}[b]{0.497\linewidth}
        \centering
        \begin{subfigure}[b]{0.511\linewidth}
        \centering
        \includegraphics[width=\linewidth, trim={12cm 0 0cm 0cm}, clip]{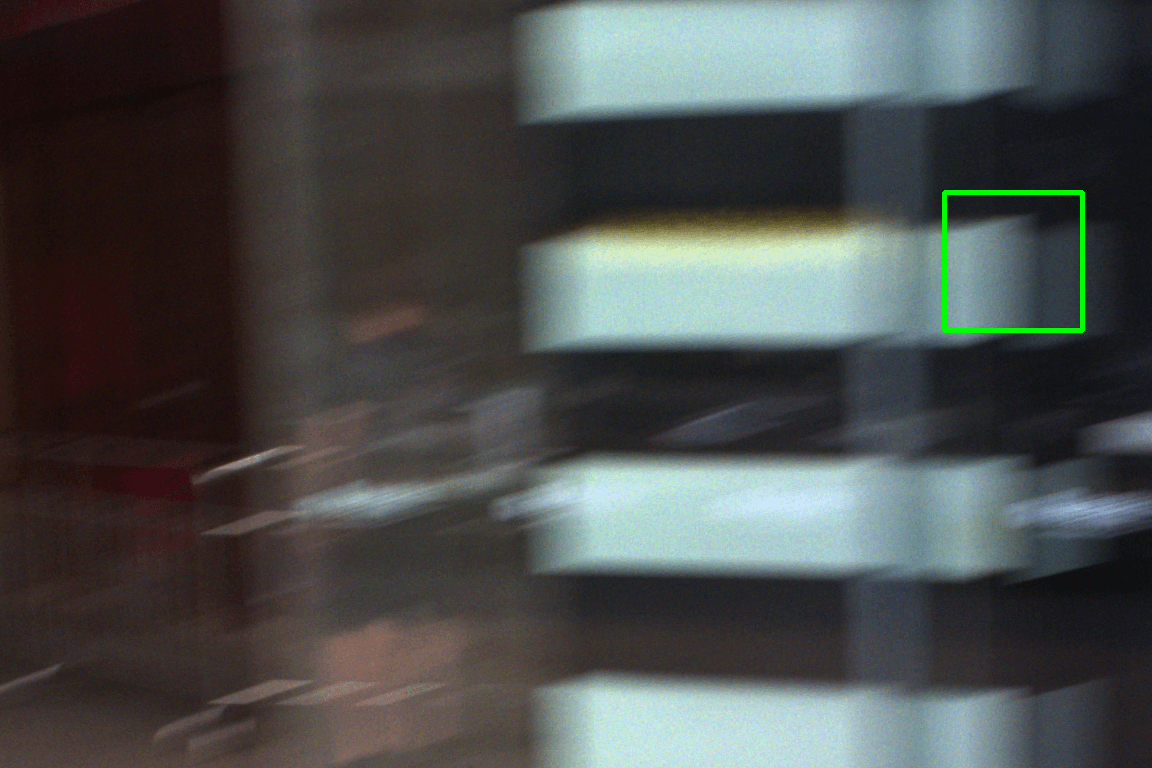}
        \vspace{-1.7em}
        \subcaption*{\centering Blurry image}
        \end{subfigure}
        \begin{subfigure}[b]{0.2195\linewidth}
        \centering
        \includegraphics[width=\linewidth, trim={0cm 0 0cm 0}]{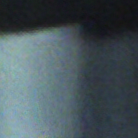}
        \vspace{-1.7em}
        \subcaption*{\centering LEVS}
        \vspace{-0.2em}
        \includegraphics[width=\linewidth, trim={0cm 0 0cm 0}]{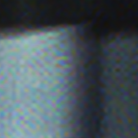}
        \vspace{-1.7em}
        \subcaption*{\centering EVDI+DASR}
        \end{subfigure}
        \begin{subfigure}[b]{0.2195\linewidth}
        \centering
        \includegraphics[width=\linewidth, trim={0cm 0 0cm 0}]{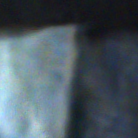}
        \vspace{-1.7em}
        \subcaption*{\centering Motion-ETR}
        \vspace{-0.2em}
        \includegraphics[width=\linewidth, trim={0cm 0 0cm 0}]{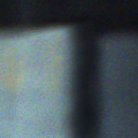}
        \vspace{-1.7em}
        \subcaption*{\centering Ours}
        \end{subfigure}
    \end{subfigure}
% % \\
% %% in 2
% \begin{subfigure}[b]{0.2215\linewidth}
% 			\includegraphics[width=\linewidth, trim={0cm 0 0cm 0}]{imgs/main/MS-RBD/in-roi-add/bbox_Blur-in.png}
% 			% \vspace{-1.7em}
% 			% \subcaption*{\centering Blurry image}
%     \end{subfigure}
%     \begin{subfigure}[b]{0.148\linewidth}
% 			\includegraphics[width=\linewidth, trim={0cm 0 0cm 0}]{imgs/main/MS-RBD/in-roi-add/Blur-in.png}
% 			% \vspace{-1.7em}
% 			% \subcaption*{\centering Blur}
%     \end{subfigure}
%     \begin{subfigure}[b]{0.148\linewidth}
% 			\includegraphics[width=\linewidth, trim={0cm 0 0cm 0}]{imgs/main/MS-RBD/in-roi-add/LEVS-in.png}
% 			% \vspace{-1.7em}
% 			% \subcaption*{\centering LEVS}
%     \end{subfigure}
%     \begin{subfigure}[b]{0.148\linewidth}
% 			\includegraphics[width=\linewidth, trim={0cm 0 0cm 0}]{imgs/main/MS-RBD/in-roi-add/Motion-ETR-in.png}
% 			% \vspace{-1.7em}
% 			% \subcaption*{\centering Motion-ETR}
%     \end{subfigure}
%     \begin{subfigure}[b]{0.148\linewidth}
% 			\includegraphics[width=\linewidth, trim={0cm 0 0cm 0}]{imgs/main/MS-RBD/in-roi-add/EVDI+DASR-in.png}
% 			% \vspace{-1.7em}
% 			% \subcaption*{\centering EVDI+DASR}
%     \end{subfigure}
%     \begin{subfigure}[b]{0.148\linewidth}
% 			\includegraphics[width=\linewidth, trim={0cm 0 0cm 0}]{imgs/main/MS-RBD/in-roi-add/Ours-in.png}
% 			% \vspace{-1.7em}
% 			% \subcaption*{\centering Ours}
%     \end{subfigure}
\\
    \vspace{-0.5em}
    \caption{Qualitative comparisons under real-world HR frames and LR events on our MS-RBD. }
    % \vspace{-0.5em}
    \label{fig:ms-rbd}
\end{figure*}

\begin{figure*}[t]
    \centering
    \begin{subfigure}[b]{0.497\linewidth}
        \centering
        \begin{subfigure}[b]{0.35\linewidth}
        \centering
        \includegraphics[width=\linewidth, trim={0cm 0 5.55cm 0cm}, clip]{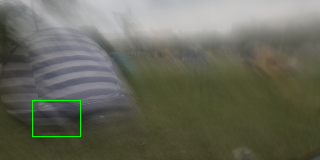}
        \vspace{-1.7em}
        \subcaption*{\centering LR blurry image}
        \end{subfigure}
        \begin{subfigure}[b]{0.2\linewidth}
        \centering
        \includegraphics[width=\linewidth, trim={0cm 0 0cm 0}]{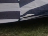}
        \vspace{-1.7em}
        \subcaption*{\centering LR GT}
        \vspace{-0.2em}
        \includegraphics[width=\linewidth, trim={0cm 0 0cm 0}]{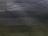}
        \vspace{-1.7em}
        \subcaption*{\centering LR blur}
        \end{subfigure}
        \begin{subfigure}[b]{0.2\linewidth}
        \centering
        \includegraphics[width=\linewidth, trim={0cm 0 0cm 0}]{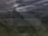}
        \vspace{-1.7em}
        \subcaption*{\centering LEVS}
        \vspace{-0.2em}
        \includegraphics[width=\linewidth, trim={0cm 0 0cm 0}]{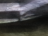}
        \vspace{-1.7em}
        \subcaption*{\centering EVDI}
        \end{subfigure}
        \begin{subfigure}[b]{0.2\linewidth}
        \centering
        \includegraphics[width=\linewidth, trim={0cm 0 0cm 0}]{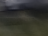}
        \vspace{-1.7em}
        \subcaption*{\centering Motion-ETR}
        \vspace{-0.2em}
        \includegraphics[width=\linewidth, trim={0cm 0 0cm 0}]{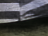}
        \vspace{-1.7em}
        \subcaption*{\centering Ours}
        \end{subfigure}
    \end{subfigure}
    %%% ---------
    \begin{subfigure}[b]{0.497\linewidth}
        \centering
        \begin{subfigure}[b]{0.35\linewidth}
        \centering
        \includegraphics[width=\linewidth, trim={5.55cm 0 0cm 0cm}, clip]{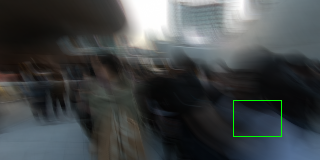}
        \vspace{-1.7em}
        \subcaption*{\centering LR blurry image}
        \end{subfigure}
        \begin{subfigure}[b]{0.2\linewidth}
        \centering
        \includegraphics[width=\linewidth, trim={0cm 0 0cm 0}]{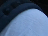}
        \vspace{-1.7em}
        \subcaption*{\centering LR GT}
        \vspace{-0.2em}
        \includegraphics[width=\linewidth, trim={0cm 0 0cm 0}]{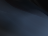}
        \vspace{-1.7em}
        \subcaption*{\centering LR blur}
        \end{subfigure}
        \begin{subfigure}[b]{0.2\linewidth}
        \centering
        \includegraphics[width=\linewidth, trim={0cm 0 0cm 0}]{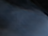}
        \vspace{-1.7em}
        \subcaption*{\centering LEVS}
        \vspace{-0.2em}
        \includegraphics[width=\linewidth, trim={0cm 0 0cm 0}]{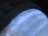}
        \vspace{-1.7em}
        \subcaption*{\centering EVDI}
        \end{subfigure}
        \begin{subfigure}[b]{0.2\linewidth}
        \centering
        \includegraphics[width=\linewidth, trim={0cm 0 0cm 0}]{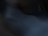}
        \vspace{-1.7em}
        \subcaption*{\centering Motion-ETR}
        \vspace{-0.2em}
        \includegraphics[width=\linewidth, trim={0cm 0 0cm 0}]{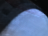}
        \vspace{-1.7em}
        \subcaption*{\centering Ours}
        \end{subfigure}
    \end{subfigure}
\\
%% HR part
\begin{subfigure}[b]{0.497\linewidth}
        \centering
        \begin{subfigure}[b]{0.35\linewidth}
        \centering
        \includegraphics[width=\linewidth, trim={0cm 0 22.2cm 0cm}, clip]{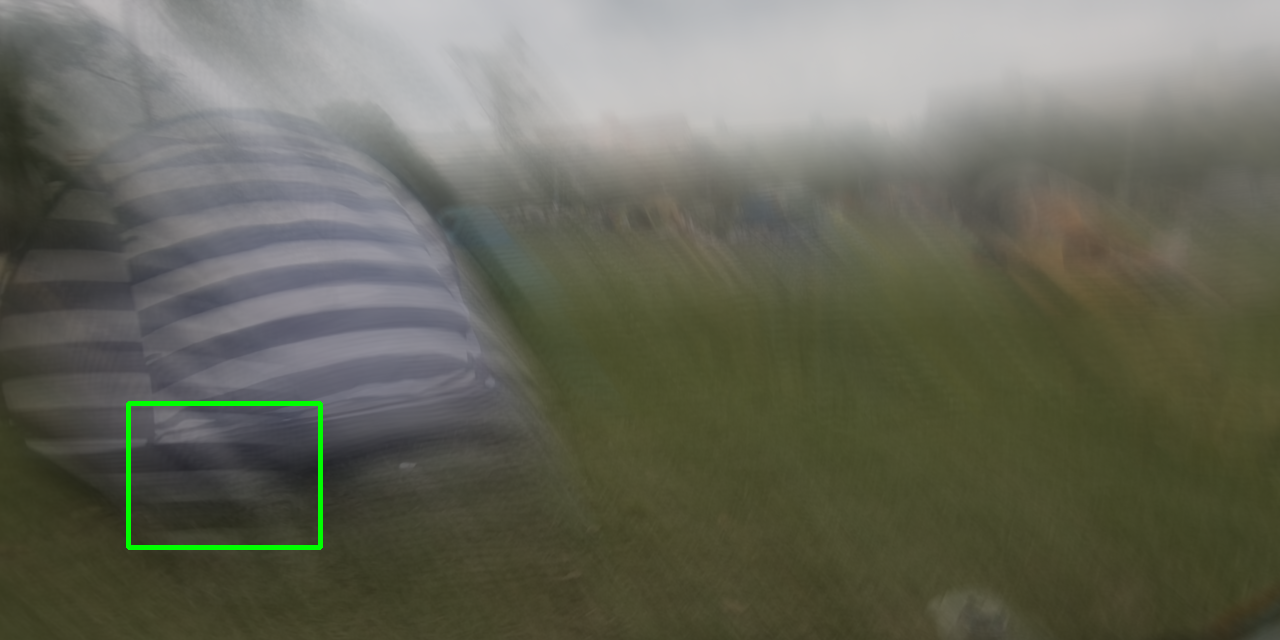}
        \vspace{-1.7em}
        \subcaption*{\centering HR blurry image}
        \end{subfigure}
        \begin{subfigure}[b]{0.2\linewidth}
        \centering
        \includegraphics[width=\linewidth, trim={0cm 0 0cm 0}]{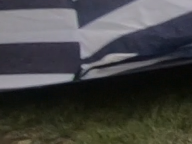}
        \vspace{-1.7em}
        \subcaption*{\centering HR GT}
        \vspace{-0.2em}
        \includegraphics[width=\linewidth, trim={0cm 0 0cm 0}]{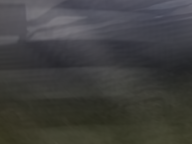}
        \vspace{-1.7em}
        \subcaption*{\centering HR blur}
        \end{subfigure}
        \begin{subfigure}[b]{0.2\linewidth}
        \centering
        \includegraphics[width=\linewidth, trim={0cm 0 0cm 0}]{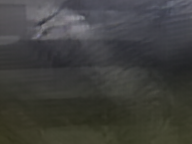}
        \vspace{-1.7em}
        \subcaption*{\centering LEVS}
        \vspace{-0.2em}
        \includegraphics[width=\linewidth, trim={0cm 0 0cm 0}]{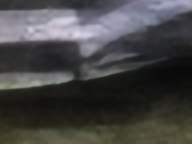}
        \vspace{-1.7em}
        \subcaption*{\centering EVDI+DASR}
        \end{subfigure}
        \begin{subfigure}[b]{0.2\linewidth}
        \centering
        \includegraphics[width=\linewidth, trim={0cm 0 0cm 0}]{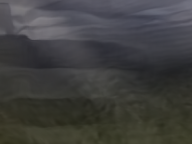}
        \vspace{-1.7em}
        \subcaption*{\centering Motion-ETR}
        \vspace{-0.2em}
        \includegraphics[width=\linewidth, trim={0cm 0 0cm 0}]{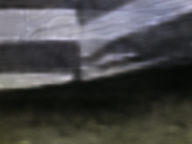}
        \vspace{-1.7em}
        \subcaption*{\centering Ours}
        \end{subfigure}
    \end{subfigure}
    %%%
\begin{subfigure}[b]{0.497\linewidth}
        \centering
        \begin{subfigure}[b]{0.35\linewidth}
        \centering
        \includegraphics[width=\linewidth, trim={22.2cm 0 0cm 0cm}, clip]{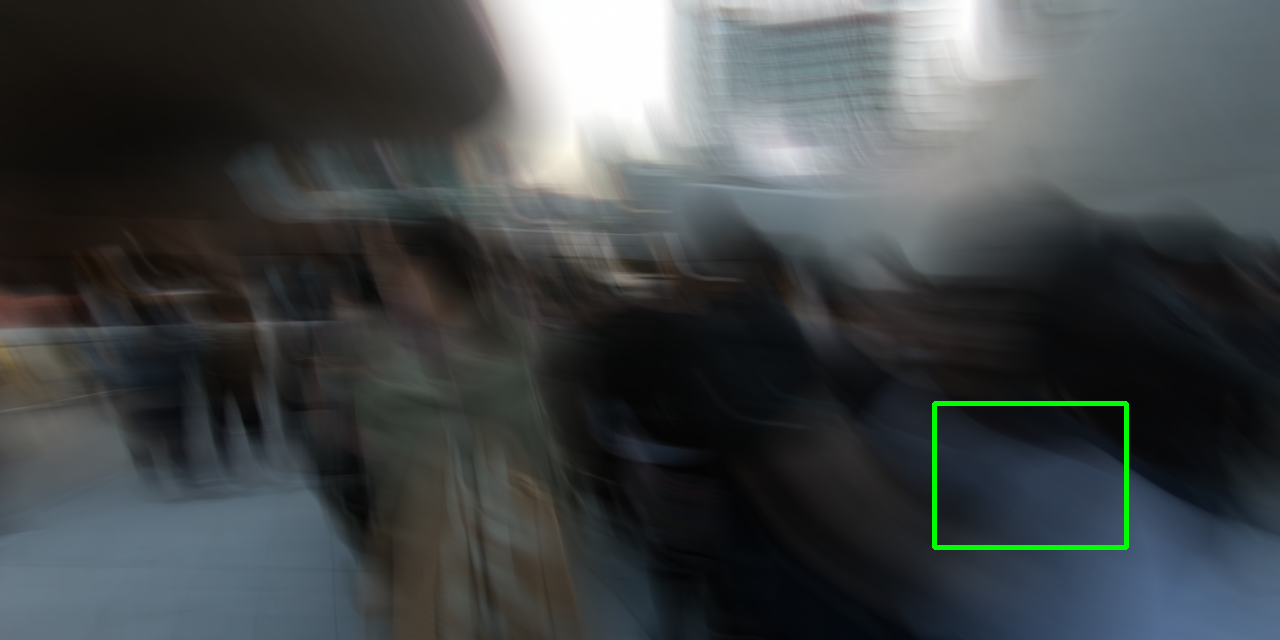}
        \vspace{-1.7em}
        \subcaption*{\centering HR blurry image}
        \end{subfigure}
        \begin{subfigure}[b]{0.2\linewidth}
        \centering
        \includegraphics[width=\linewidth, trim={0cm 0 0cm 0}]{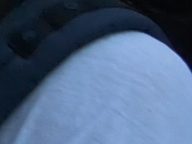}
        \vspace{-1.7em}
        \subcaption*{\centering HR GT}
        \vspace{-0.2em}
        \includegraphics[width=\linewidth, trim={0cm 0 0cm 0}]{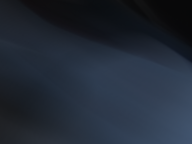}
        \vspace{-1.7em}
        \subcaption*{\centering HR blur}
        \end{subfigure}
        \begin{subfigure}[b]{0.2\linewidth}
        \centering
        \includegraphics[width=\linewidth, trim={0cm 0 0cm 0}]{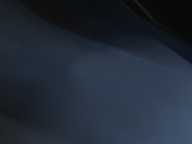}
        \vspace{-1.7em}
        \subcaption*{\centering LEVS}
        \vspace{-0.2em}
        \includegraphics[width=\linewidth, trim={0cm 0 0cm 0}]{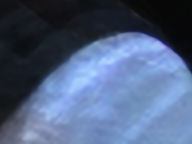}
        \vspace{-1.7em}
        \subcaption*{\centering EVDI+DASR}
        \end{subfigure}
        \begin{subfigure}[b]{0.2\linewidth}
        \centering
        \includegraphics[width=\linewidth, trim={0cm 0 0cm 0}]{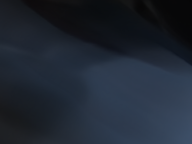}
        \vspace{-1.7em}
        \subcaption*{\centering Motion-ETR}
        \vspace{-0.2em}
        \includegraphics[width=\linewidth, trim={0cm 0 0cm 0}]{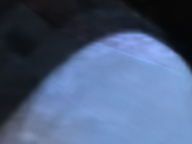}
        \vspace{-1.7em}
        \subcaption*{\centering Ours}
        \end{subfigure}
    \end{subfigure}
    % \vspace{-1.5em}
	\caption{ Qualitative comparisons under different spatial scales $\mathcal{R}(B_T,\mathcal{E}_\mathcal{T})=1$ (LR blur, top row) and $\mathcal{R}(B_T,\mathcal{E}_\mathcal{T})=4$ (HR blur, bottom row) on the Ev-REDS dataset. GT indicates ground-truth images.
 % Our results are converted to gray-scale for fair comparisons. 
% 	Details are zoomed in for a better view.
	}
    % \vspace{-0.5em}
	\label{fig:ev-reds}
\end{figure*}
\begin{table*}[!h]
\centering
\small
\renewcommand{\arraystretch}{1.2}
\caption{Quantitative comparisons under different spatial scales ($\mathcal{R}(B_T,\mathcal{E}_\mathcal{T})=1$ and $\mathcal{R}(B_T,\mathcal{E}_\mathcal{T})=4$) on the Ev-REDS dataset. 
Image (DASR), video (RealBasicVSR), and event (EventZoom) super-resolution techniques are employed to assist event-based deblurring methods in the case of $\mathcal{R}(B_T,\mathcal{E}_\mathcal{T})=4$.
Symbol $/$ denotes unavailable metrics as some methods only work with gray images. For those that work with color images (LEVS, Motion-ETR, EVDI, and ours), their results are also converted to gray-scale for computing gray metrics. Best and second-best results are \textbf{bolded} and \underline{underlined}, respectively.
% The color metrics of some algorithms are not available since they only work with gray images, and we convert the color results to gray-scale for computing gray metrics.
% For the methods that work with color images, their results are also converted to gray-scale for computing gray metrics. 
}
% \vspace{-0.5em}
\begin{tabular}{lccccccccccc}
\toprule[2pt]
\multirow{3}{*}{Method} & \multicolumn{5}{c}{Comparison under \textcolor{black}{$\mathcal{R}(B_T,\mathcal{E}_\mathcal{T})=1$}}                            &  & \multicolumn{5}{c}{Comparison under \textcolor{black}{$\mathcal{R}(B_T,\mathcal{E}_\mathcal{T})=4$}}                            \\ \cline{2-6} \cline{8-12} 
                        & \multicolumn{2}{c}{Color metric} &  & \multicolumn{2}{c}{Gray metric} &  & \multicolumn{2}{c}{Color metric} &  & \multicolumn{2}{c}{Gray metric} \\ \cline{2-3} \cline{5-6} \cline{8-9} \cline{11-12} 
                        & PSNR $\uparrow$       & SSIM $\uparrow$       &  & PSNR $\uparrow$      & SSIM $\uparrow$       &  & PSNR $\uparrow$       & SSIM $\uparrow$       &  & PSNR $\uparrow$      & SSIM $\uparrow$       \\ \midrule[1pt]
LEVS \cite{jin2018learning}                   & 18.24       & 0.4665      &  & 18.36      & 0.4680      &  & 18.62       & 0.4612      &  & 18.75      & 0.4644      \\
Motion-ETR \cite{zhang2021exposure}            & 17.79       & 0.4376      &  & 17.90      & 0.4388      &  & 18.23       & 0.4292      &  & 18.34      & 0.4320      \\
EDI \cite{edi_pan2019bringing} (+DASR \cite{wang2021unsupervised})             & /           & /           &  & 20.41      & 0.6067      &  & /           & /           &  & 18.81      & 0.4553      \\
eSL-Net \cite{esl_wang2020event}                & /           & /           &  & 19.41      & 0.7119      &  & /           & /           &  & 18.96      & 0.5604      \\
RED \cite{red_xu2021motion} (+DASR \cite{wang2021unsupervised})             & /           & /           &  & \textcolor{black}{23.21}      & \textcolor{black}{\underline{0.7959}}      &  & /           & /           &  & 22.60      & 0.6350      \\
EVDI \cite{zhang2022unifying} (+EventZoom \cite{duan2021eventzoom})      & \underline{23.88}       & \underline{0.7789}      &  & \underline{24.37}      & 0.7917           &  & 18.93       & 0.4815      &  & 19.07      & 0.4848      \\
EVDI \cite{zhang2022unifying} (+RealBasicVSR \cite{chan2022investigating})      & \underline{23.88}       & \underline{0.7789}      &  & \underline{24.37}      & 0.7917           &  & 23.33       & \underline{0.6441}      &  & 23.79      & \underline{0.6568}      \\
EVDI \cite{zhang2022unifying} (+DASR \cite{wang2021unsupervised})            & \underline{23.88}       & \underline{0.7789}      &  & \underline{24.37}      & 0.7917      &  & \underline{23.35}       & 0.6368      &  & \underline{23.83}      & 0.6477      \\ \midrule[1pt]
Ours                    & \textbf{24.12}       & \textbf{0.7898}      &  & \textbf{24.63}      & \textbf{0.8022}      &  & \textbf{23.95}       & \textbf{0.6647}      &  & \textbf{24.43}      & \textbf{0.6749}      \\ \bottomrule[2pt]
\end{tabular}
% \vspace{-1em}
\label{tab:ev-reds-main}
\end{table*}

% 23.96      & \textbf{0.8525}
\noindent \textbf{Temporal and Spatial Generalization.} The second stage of training aims to generalize the deblurring performance of SAN in both temporal and spatial dimensions. For temporal generalization, we propose a self-distillation loss
\begin{equation}\label{eq:loss_tg}
    \mathcal{L}_{TG} = \|\overline{\operatorname{SAN}}([t,t];B_T,\mathcal{E}_{\mathcal{T}}) - \operatorname{SAN}([t,t];B_{\tilde{T}}, \mathcal{E}_{\tilde{\mathcal{T}}})  \|_1,
\end{equation}
where $\overline{\operatorname{SAN}}$ indicates a fixed teacher model pre-trained using $\mathcal{L}_{BC}$ and $\mathcal{L}_{SC}$, and $\operatorname{SAN}$ is the student network loaded from $\overline{\operatorname{SAN}}$ and continuing to train. 
Since $\overline{\operatorname{SAN}}$ can recover relatively more reliable latent images from the less blurred frame $B_T$, we treat the output of $\overline{\operatorname{SAN}}$ as pseudo-ground-truth images and teach $\operatorname{SAN}$ to deblur the more blurred frame $B_{\tilde{T}}$, which improves the deblurring ability of SAN and generalizes its performance to handle different temporal scales of motion blur.

% To further generalize the deblurring performance of SAN in the spatial domain, \ie, from $\mathcal{R}(B_T,\mathcal{E}_\mathcal{T}) = \bar{R}$ to $\mathcal{R}(B_T,\mathcal{E}_\mathcal{T}) \in \mathbf{R}^{*}(\bar{R})$,

\par 
% spatial generalization, motivation
% With the above constraints, our SAN learns to deblur HR frames with LR events under a fixed spatial ratio $\mathcal{R}(B_T,\mathcal{E}_\mathcal{T}) = \bar{R}$, but it suffers from inconsistent performance when inferring data with different spatial scales, \eg, $\mathcal{R}(B_T,\mathcal{E}_\mathcal{T}) \in \mathbf{R}^{*}(\bar{R})$. 
With the above constraints, our SAN learns to deblur HR frames with LR events at a fixed spatial ratio $\mathcal{R}(B_T,\mathcal{E}_\mathcal{T}) = \bar{R}$, but its performance in handling different spatial scales of motion blur, \ie, $\mathcal{R}(B_T,\mathcal{E}_\mathcal{T}) \in \mathbf{R}^{*}(\bar{R})$, is not guaranteed. 
% suffers from inconsistent performance when inferring data with different spatial scales, \eg, $\mathcal{R}(B_T,\mathcal{E}_\mathcal{T}) \in \mathbf{R}^{*}(\bar{R})$. 
To generalize the deblurring performance in the spatial domain, we encourage SAN to adaptively project event features according to the input spatial scale $\mathcal{R}(B_T,\mathcal{E}_\mathcal{T})$. 
% Based on $\mathcal{L}_{TG}$, we form inputs with varying spatial scales by randomly down-sampling $B_{\tilde{T}}$ to $B_{\tilde{T}}^{\downarrow}$ with $\forall \mathcal{R}(B_{\tilde{T}}^{\downarrow},\mathcal{E}_\mathcal{T})\in[1, \bar{R})$, and then formulate the constraint as
Specifically, we first form inputs with varying spatial scales by randomly down-sampling $B_{\tilde{T}}$ to $B_{\tilde{T}}^{\downarrow}$ with $\forall \mathcal{R}(B_{\tilde{T}}^{\downarrow},\mathcal{E}_{\tilde{\mathcal{T}}})\in[1, \bar{R})$, and then formulate the constraint based on the idea of self-distillation,
\begin{equation}\label{eq:loss_SG}
    \mathcal{L}_{SG} = \|\overline{\operatorname{SAN}}^{\downarrow}([t,t];B_T,\mathcal{E}_{\mathcal{T}}) - \operatorname{SAN}([t,t];B^{\downarrow}_{\tilde{T}}, \mathcal{E}_{\tilde{\mathcal{T}}})  \|_1,
\end{equation}
where $\overline{\operatorname{SAN}}^{\downarrow}$ means $\overline{\operatorname{SAN}}$ followed by a down-sampling operation. With $\mathcal{L}_{SG}$, our SAN is able to propagate the deblurring performance under $\mathcal{R}(B_T,\mathcal{E}_\mathcal{T}) = \bar{R}$ to different input spatial scales $\mathcal{R}(B_T,\mathcal{E}_\mathcal{T}) \in \mathbf{R}^{*}(\bar{R})$.

% learn the distribution differences with respect to $\mathcal{R}(B_{\tilde{T}}^{\downarrow},\mathcal{E}_\mathcal{T})$ and generalize the deblurring performance to inputs at different spatial scales.

% our SAN is able to propagate its deblurring performance to inputs at different spatial scales.

% To generalize the deblurring performance in the spatial domain, we encourage SAN to adaptively project events to fit different spatial scales of blurry frames.
% % according to different input spatial scales. 
% % Considering that the principles of event up-sampling remain unclear due to the varying distributions at different spatial scales \cite{gehrig2022high}, we opt to simulate different $\mathcal{R}(B_T,\mathcal{E}_\mathcal{T})$ by image down-sampling. 
% Specifically, we first randomly down-sample $B_{\tilde{T}}$ to $B_{\tilde{T}}^{\downarrow}$ 
% with $\forall \mathcal{R}(B_{\tilde{T}}^{\downarrow},\mathcal{E}_\mathcal{T})\in[1, \bar{R})$, and then formulate the constraint as
% % \begin{equation}\label{eq:loss_sg}
% $$
%     \mathcal{L}_{SG} = \|\overline{\operatorname{SAN}}^{\downarrow}([t,t];B_T,\mathcal{E}_{\mathcal{T}}) - \operatorname{SAN}([t,t];B^{\downarrow}_{\tilde{T}}, \mathcal{E}_{\tilde{\mathcal{T}}})  \|_1,
% $$
% % \end{equation}
% where $\overline{\operatorname{SAN}}^{\downarrow}$ means $\overline{\operatorname{SAN}}$ followed by a down-sampling operation. With $\mathcal{L}_{SG}$, our SAN is able to propagate its deblurring performance to inputs at different spatial scales.

\par 
Finally, our self-supervised learning framework can be summarized as 
\begin{equation}\label{eq:loss_all}
    \mathcal{L} = \beta_{BC} \mathcal{L}_{BC} +  \beta_{SC} \mathcal{L}_{SC} +  \beta_{TG} \mathcal{L}_{TG} + \beta_{SG}\mathcal{L}_{SG},
\end{equation}
where  $\beta_{BC}, \beta_{SC}, \beta_{TG}, \beta_{SG}$ indicate the balancing parameters.
Compared with the previous self-supervised EVDI \cite{zhang2022unifying}, our method generalizes the deblurring performance to handle the varying blurriness levels and different spatial scales of real motion blur. Furthermore, our approach shows better efficiency by design. For example, EVDI supervises brightness consistency via reblurring techniques, which require restoring a large number (49 in EVDI) of latent images per input during training, while ours efficiently fulfills this by learning blur2blur conversion (please see the supplementary material for detailed comparisons).

\begin{figure*}[t]
    \centering
    \begin{subfigure}[b]{0.497\linewidth}
        \centering
        \begin{subfigure}[b]{0.371\linewidth}
        \centering
        \includegraphics[width=\linewidth, trim={0cm 0 0cm 0cm}, clip]{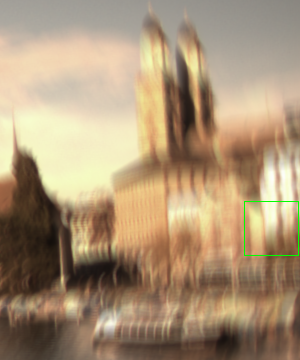}
        \vspace{-1.7em}
        \subcaption*{\centering Image with normal blur}
        \end{subfigure}
        \begin{subfigure}[b]{0.2\linewidth}
        \centering
        \includegraphics[width=\linewidth, trim={0cm 0 0cm 0}]{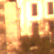}
        \vspace{-1.7em}
        \subcaption*{\centering GT}
        \vspace{-0.2em}
        \includegraphics[width=\linewidth, trim={0cm 0 0cm 0}]{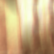}
        \vspace{-1.7em}
        \subcaption*{\centering Normal blur}
        \end{subfigure}
        \begin{subfigure}[b]{0.2\linewidth}
        \centering
        \includegraphics[width=\linewidth, trim={0cm 0 0cm 0}]{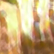}
        \vspace{-1.7em}
        \subcaption*{\centering LEVS}
        \vspace{-0.2em}
        \includegraphics[width=\linewidth, trim={0cm 0 0cm 0}]{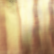}
        \vspace{-1.7em}
        \subcaption*{\centering EVDI}
        \end{subfigure}
        \begin{subfigure}[b]{0.2\linewidth}
        \centering
        \includegraphics[width=\linewidth, trim={0cm 0 0cm 0}]{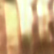}
        \vspace{-1.7em}
        \subcaption*{\centering Motion-ETR}
        \vspace{-0.2em}
        \includegraphics[width=\linewidth, trim={0cm 0 0cm 0}]{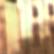}
        \vspace{-1.7em}
        \subcaption*{\centering Ours}
        \end{subfigure}
    \end{subfigure}
    %%% ---------
    \begin{subfigure}[b]{0.497\linewidth}
        \centering
        \begin{subfigure}[b]{0.371\linewidth}
        \centering
        \includegraphics[width=\linewidth, trim={0cm 0 0cm 0cm}, clip]{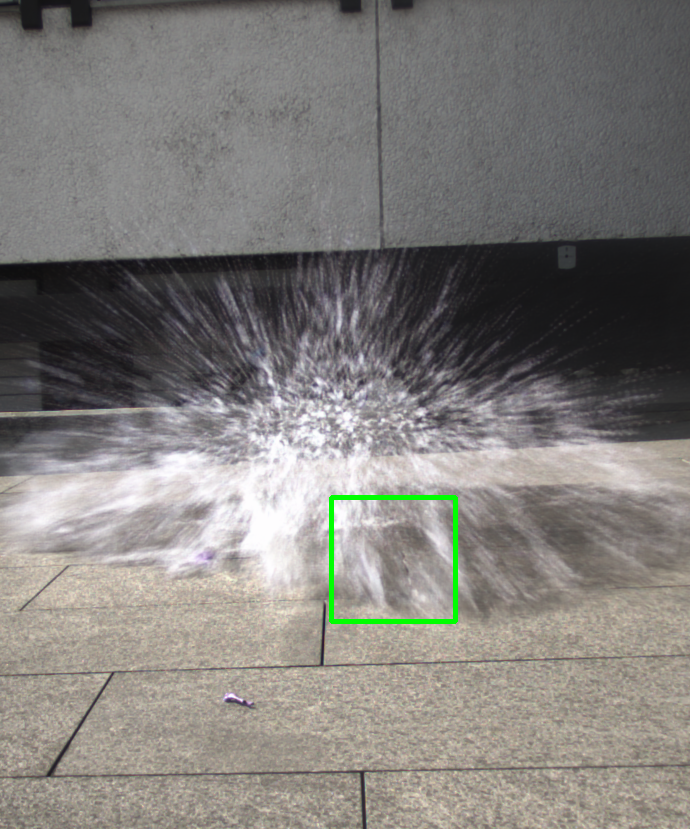}
        \vspace{-1.7em}
        \subcaption*{\centering Image with normal blur}
        \end{subfigure}
        \begin{subfigure}[b]{0.2\linewidth}
        \centering
        \includegraphics[width=\linewidth, trim={0cm 0 0cm 0}]{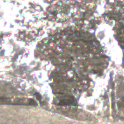}
        \vspace{-1.7em}
        \subcaption*{\centering GT}
        \vspace{-0.2em}
        \includegraphics[width=\linewidth, trim={0cm 0 0cm 0}]{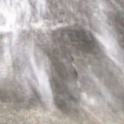}
        \vspace{-1.7em}
        \subcaption*{\centering Normal blur}
        \end{subfigure}
        \begin{subfigure}[b]{0.2\linewidth}
        \centering
        \includegraphics[width=\linewidth, trim={0cm 0 0cm 0}]{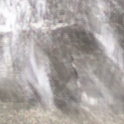}
        \vspace{-1.7em}
        \subcaption*{\centering LEVS}
        \vspace{-0.2em}
        \includegraphics[width=\linewidth, trim={0cm 0 0cm 0}]{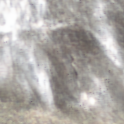}
        \vspace{-1.7em}
        \subcaption*{\centering EVDI}
        \end{subfigure}
        \begin{subfigure}[b]{0.2\linewidth}
        \centering
        \includegraphics[width=\linewidth, trim={0cm 0 0cm 0}]{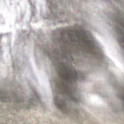}
        \vspace{-1.7em}
        \subcaption*{\centering Motion-ETR}
        \vspace{-0.2em}
        \includegraphics[width=\linewidth, trim={0cm 0 0cm 0}]{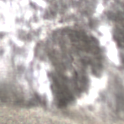}
        \vspace{-1.7em}
        \subcaption*{\centering Ours}
        \end{subfigure}
    \end{subfigure}
\\ % -------------------------
\begin{subfigure}[b]{0.497\linewidth}
        \centering
        \begin{subfigure}[b]{0.371\linewidth}
        \centering
        \includegraphics[width=\linewidth, trim={0cm 0 0cm 0cm}, clip]{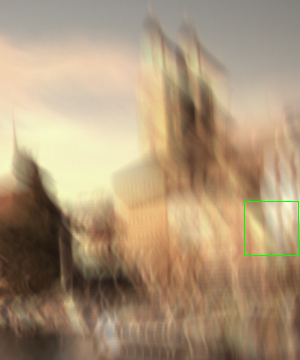}
        \vspace{-1.7em}
        \subcaption*{\centering Image with large blur}
        \end{subfigure}
        \begin{subfigure}[b]{0.2\linewidth}
        \centering
        \includegraphics[width=\linewidth, trim={0cm 0 0cm 0}]{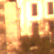}
        \vspace{-1.7em}
        \subcaption*{\centering GT}
        \vspace{-0.2em}
        \includegraphics[width=\linewidth, trim={0cm 0 0cm 0}]{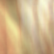}
        \vspace{-1.7em}
        \subcaption*{\centering Large blur}
        \end{subfigure}
        \begin{subfigure}[b]{0.2\linewidth}
        \centering
        \includegraphics[width=\linewidth, trim={0cm 0 0cm 0}]{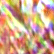}
        \vspace{-1.7em}
        \subcaption*{\centering LEVS}
        \vspace{-0.2em}
        \includegraphics[width=\linewidth, trim={0cm 0 0cm 0}]{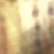}
        \vspace{-1.7em}
        \subcaption*{\centering EVDI}
        \end{subfigure}
        \begin{subfigure}[b]{0.2\linewidth}
        \centering
        \includegraphics[width=\linewidth, trim={0cm 0 0cm 0}]{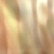}
        \vspace{-1.7em}
        \subcaption*{\centering Motion-ETR}
        \vspace{-0.2em}
        \includegraphics[width=\linewidth, trim={0cm 0 0cm 0}]{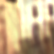}
        \vspace{-1.7em}
        \subcaption*{\centering Ours}
        \end{subfigure}
    \end{subfigure}
    %%% ---------
    \begin{subfigure}[b]{0.497\linewidth}
        \centering
        \begin{subfigure}[b]{0.371\linewidth}
        \centering
        \includegraphics[width=\linewidth, trim={0cm 0 0cm 0cm}, clip]{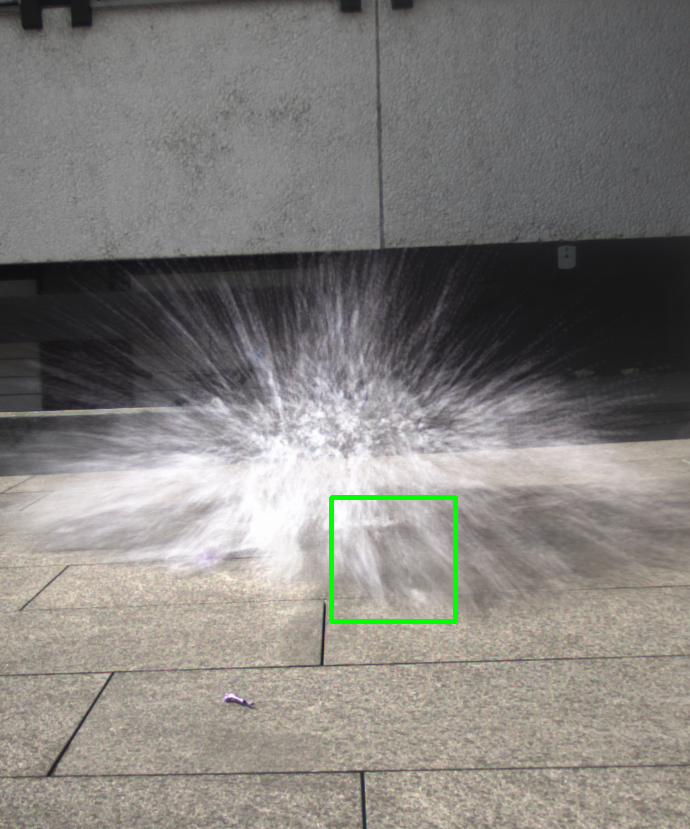}
        \vspace{-1.7em}
        \subcaption*{\centering Image with large blur}
        \end{subfigure}
        \begin{subfigure}[b]{0.2\linewidth}
        \centering
        \includegraphics[width=\linewidth, trim={0cm 0 0cm 0}]{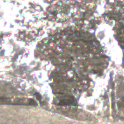}
        \vspace{-1.7em}
        \subcaption*{\centering GT}
        \vspace{-0.2em}
        \includegraphics[width=\linewidth, trim={0cm 0 0cm 0}]{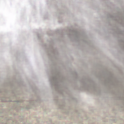}
        \vspace{-1.7em}
        \subcaption*{\centering Large blur}
        \end{subfigure}
        \begin{subfigure}[b]{0.2\linewidth}
        \centering
        \includegraphics[width=\linewidth, trim={0cm 0 0cm 0}]{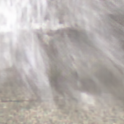}
        \vspace{-1.7em}
        \subcaption*{\centering LEVS}
        \vspace{-0.2em}
        \includegraphics[width=\linewidth, trim={0cm 0 0cm 0}]{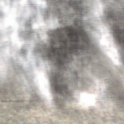}
        \vspace{-1.7em}
        \subcaption*{\centering EVDI}
        \end{subfigure}
        \begin{subfigure}[b]{0.2\linewidth}
        \centering
        \includegraphics[width=\linewidth, trim={0cm 0 0cm 0}]{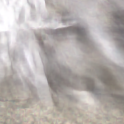}
        \vspace{-1.7em}
        \subcaption*{\centering Motion-ETR}
        \vspace{-0.2em}
        \includegraphics[width=\linewidth, trim={0cm 0 0cm 0}]{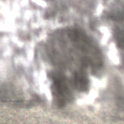}
        \vspace{-1.7em}
        \subcaption*{\centering Ours}
        \end{subfigure}
    \end{subfigure}
    % \vspace{-1.5em}
	\caption{ Qualitative comparisons under normal blur (top row) and large blur (bottom row) on the HS-ERGB dataset.
	}
    % \vspace{-0.5em}
	\label{fig:hs-ergb}
\end{figure*}
\begin{table*}[th]
\centering
\small
\renewcommand{\arraystretch}{1.2}
\caption{Quantitative comparisons under $\mathcal{R}(B_T,\mathcal{E}_{\mathcal{T}})=1$ and different temporal scales (normal and large blur) on the HS-ERGB dataset.  
The symbol $/$ denotes unavailable metrics as some algorithms only work with gray images. The results of color models (LEVS, Motion-ETR, EVDI, and ours) are converted to gray-scale for computing gray metrics. 
% For the methods that work with color images (LEVS, Motion-ETR, EVDI, and ours), their results are also converted to gray-scale for computing gray metrics. 
% For the methods that work with color images, their results are also converted to gray-scale for computing gray metrics.
}
% \vspace{-0.5em}
\begin{tabular}{lccccccccccc}
\toprule[2pt]
\multirow{3}{*}{Method} & \multicolumn{5}{c}{Comparison under normal blur}                            &  & \multicolumn{5}{c}{Comparison under large blur}                            \\ \cline{2-6} \cline{8-12} 
                        & \multicolumn{2}{c}{Color metric} &  & \multicolumn{2}{c}{Gray metric} &  & \multicolumn{2}{c}{Color metric} &  & \multicolumn{2}{c}{Gray metric} \\ \cline{2-3} \cline{5-6} \cline{8-9} \cline{11-12} 
                        & PSNR $\uparrow$       & SSIM $\uparrow$       &  & PSNR $\uparrow$      & SSIM $\uparrow$       &  & PSNR $\uparrow$       & SSIM $\uparrow$       &  & PSNR $\uparrow$      & SSIM $\uparrow$       \\ \midrule[1pt]
LEVS \cite{jin2018learning}                    & 22.13       & 0.5548      &     & 22.70      & 0.5935      &  & 21.72       & 0.5429      &     & 22.06      & 0.5741      \\
Motion-ETR \cite{zhang2021exposure}              & 23.79       & 0.6276      &     & 24.05      & 0.6464      &  & 22.73       & 0.5842      &     & 22.88      & 0.6010      \\
EDI \cite{edi_pan2019bringing}                    & /           & /           &     & 23.93      & 0.7043      &  & /           & /           &     & 22.33      & 0.6517      \\
eSL-Net \cite{esl_wang2020event}                & /           & /           &     & 24.10      & 0.6811      &  & /           & /           &     & 22.76      & 0.6248      \\
RED \cite{red_xu2021motion}                    & /           & /           &     & \underline{26.05}      & 0.7234      &  & /           & /           &     & \underline{24.81}      & 0.6676      \\
EVDI \cite{zhang2022unifying}                   & \underline{25.13}       & \underline{0.7072}      &     & 25.49      & \underline{0.7312}      &  & \underline{24.08}       & \underline{0.6637}      &     & 24.35      & \underline{0.6856}      \\ \midrule[1pt]
Ours                    & \textbf{26.22}       & \textbf{0.7292}      &     & \textcolor{black}{\textbf{26.87}}      & \textcolor{black}{\textbf{0.7529}}      &  & \textbf{25.41}       & \textbf{0.6936}      &     & \textbf{25.94}      & \textbf{0.7168}      \\ \bottomrule[2pt]
\end{tabular}
% \vspace{-1.5em}
\label{tab:ev-hsergb-main}
\end{table*}

\section{Experiments and Analysis}
\subsection{Experimental Setup}
% dataset, details
\noindent \textbf{Datasets.} Three different datasets containing synthetic, semi-synthetic, and real-world blurry frames and events are employed in our experiments for evaluation. 
\par 
\textbf{Ev-REDS:} We build a synthetic dataset upon REDS \cite{gopro_nah2019ntire} for evaluation on different spatial scales. We first crop the sharp images to size $1280\times640$ and down-sample them to $320\times160$ to form HR and LR sequences. For each sequence, we generate high frame-rate videos by interpolating $7$ images between consecutive frames using RIFE \cite{huang2022real}, and then synthesize blurry frames by averaging 49 sharp images of the high frame-rate videos. Events are generated via VID2E \cite{gehrig2020video} on the LR sequences to form two sets with different spatial scales $\mathcal{R}(B_T,\mathcal{E}_{\mathcal{T}})=4$ (HR frames and LR events, for training and testing) and $\mathcal{R}(B_T,\mathcal{E}_{\mathcal{T}})=1$ (LR frames and events, only for testing).
% We employ VID2E \cite{gehrig2020video} to generate events on the LR sequences to form the data of $\mathcal{R}(B_T,\mathcal{E}_{\mathcal{T}})=4$ (HR frames and LR events, for training and testing) and $\mathcal{R}(B_T,\mathcal{E}_{\mathcal{T}})=1$ (LR frames and events, only for testing).

\par 
\textbf{HS-ERGB:} HS-ERGB dataset \cite{tulyakov2021time} contains sharp videos and real events at the same spatial resolution, and thus we employ it for evaluation on different temporal scales of motion blur. We first increase the frame rate of the original videos by interpolating $7$ images between consecutive frames with Time Lens \cite{tulyakov2021time}, and then synthesize two types of blurry videos by averaging 49 and 97 frames, which we call normal and large blur, respectively. The set with normal blur is used for training and testing, and the one with large blur is only used for testing.

% Since the original dataset contains many static frames, we manually filter them and leave only dynamic scenes in our experiments.

% We select 4 sequences for testing and leave others for training

\par 
\textbf{MS-RBD:} Due to the lack of available real-world datasets with HR blurry frames and LR events, we construct a Multi-Scale Real-world Blurry Dataset (MS-RBD) with a FLIR Blackfly S global shutter RGB camera and a DAVIS346 camera. A beam splitter is implemented in front of the two cameras with $50\%$ splitting. In total, we collect 32 sequences of data composed of 22 indoor and 10 outdoor scenes, where the blur caused by camera ego-motion and dynamic scenes are both considered. We also set the frame rate of the FLIR camera to 30 and 15 FPS to imitate the blur at different temporal scales. 
After spatial alignment, each sequence contains 60 RGB frames at size $1152\times 768$ and the corresponding $288\times 192$ events. More details can be found in the supplementary material.

\par 
\noindent \textbf{Implementation Details.} Our SAN is implemented in the Pytorch platform and trained on NVIDIA GeForce RTX 2080 Ti GPUs with batch size 3. We set the number of temporal bins $N=16$ and the temporal scale parameter $M=2$. The Adam optimizer \cite{kingma2014adam} and the SGDR schedule \cite{loshchilov2016sgdr} are employed for training.
% and randomly crop the images to $256 \times 256$ patches for training, where the Adam optimizer \cite{kingma2014adam} and the SGDR schedule \cite{loshchilov2016sgdr} are employed. 
We first train an SAN with the parameters $[\beta_{BC}, \beta_{SC}, \beta_{TG},\beta_{SG}]=[50,1,0,0]$ and learning rate $1\times10^{-3}$ for 210 epochs. With the pre-trained $\overline{\operatorname{SAN}}$ as the teacher model, we continue training the SAN with $[\beta_{BC}, \beta_{SC}, \beta_{TG},\beta_{SG}]=[50,1,50,50]$ and learning rate $5\times10^{-4}$ for 15 cycles. Every cycle lasts for 30 epochs, and we update the teacher model at the end of each cycle.

\subsection{Benchmarking}
We evaluate the proposed method by comparing with the state-of-the-art deblurring approaches, including frame-based algorithms \textcolor{black}{ LEVS \cite{jin2018learning}, Motion-ETR \cite{zhang2021exposure}, and event-based methods EDI \cite{edi_pan2019bringing}, eSL-Net \cite{esl_wang2020event}, RED \cite{red_xu2021motion}, and EVDI \cite{zhang2022unifying}.} Since we assume real-world scenarios without available ground-truth images, only the self-supervised EVDI can be trained under such circumstances, and we use the official codes for re-training. In the case of $\mathcal{R}(B_T,\mathcal{E}_{\mathcal{T}})>1$, we employ state-of-the-art image, video, and event super-resolution techniques DASR \cite{wang2021unsupervised}, RealBasicVSR \cite{chan2022investigating}, and EventZoom \cite{duan2021eventzoom} to assist event-based deblurring methods as they only accept inputs of $\mathcal{R}(B_T,\mathcal{E}_{\mathcal{T}})=1$. Metrics PSNR and SSIM \cite{wang2003multiscale} are computed based on sequence restoration, \ie, restoring 7 sharp images from one blurry input, for quantitative evaluation.
% to evaluate both the reconstruction performance and the 
% employed for quantitative evaluation .

% \textcolor{red}{exp setup: single / sequence}
% Metrics PSNR and SSIM \cite{wang2003multiscale} are employed for quantitative evaluation on both color and gray domains.
% \vspace{-0.2em}
\par 
Tab.~\ref{tab:ev-reds-main} validates the robust performance of our proposed approach under different spatial scales. Although frame-based methods can directly process blurry frames at different spatial resolutions, they often fail in highly dynamic scenes with complex motions because of motion ambiguity, as depicted in Fig.~\ref{fig:ms-rbd}. For event-based algorithms, eSL-Net is able to produce HR results by simultaneously considering motion deblurring and image super-resolution. However, eSL-Net only receives blurry frames and events of the same spatial resolution, which limits its performance due to the information loss caused by image down-sampling. Similar to eSL-Net, previous event-based methods generally assume $\mathcal{R}(B_T,\mathcal{E}_{\mathcal{T}})=1$ for inputs, and thus super-resolution techniques are necessary to restore HR results. As shown in Tab.~\ref{tab:ev-reds-main} and Fig.~\ref{fig:ev-reds}, such cascaded scheme often leads to sub-optimal performance as the deblurring or super-resolution errors will be propagated to the subsequent stage. 

\begin{table}[t]
\centering
\small
\renewcommand{\arraystretch}{1.2}
\caption{Ablation study of our self-supervised learning framework under different spatial scales $\mathcal{R}(B_T,\mathcal{E}_{\mathcal{T}})=1$ (LR) and $\mathcal{R}(B_T,\mathcal{E}_{\mathcal{T}})=4$ (HR) on the Ev-REDS dataset.}
% \vspace{-0.5em}
\begin{tabular}{c|cccc|c}
\toprule[2pt]
ID & $\mathcal{L}_{BC}$        & $\mathcal{L}_{SC}$      & $\mathcal{L}_{TG}$       & $\mathcal{L}_{SG}$        &    LR /  HR PSNR $\uparrow$           \\ \midrule[1pt]
\#1  & \checkmark &           &           &           & 19.20 / 19.95  \\
\#2  &           & \checkmark &           &           & 18.94 / 18.95   \\
\#3  & \checkmark & \checkmark &           &           & 21.77  / {23.39}  \\
\#4  & \checkmark & \checkmark & \checkmark &           & {21.73} / \textbf{24.00}  \\
\#5  & \checkmark & \checkmark &  & \checkmark          & \underline{23.46} / 23.23  \\
\#6  & \checkmark & \checkmark & \checkmark & \checkmark & \textbf{24.12} / \underline{23.95}  \\ \bottomrule[2pt]
\end{tabular}
% \vspace{-1.5em}
\label{tab:ablation}
\end{table}

% \begin{table}[t]
% \centering
% \footnotesize
% \renewcommand{\arraystretch}{1.1}
% \caption{Ablation study of our proposed self-supervised learning framework under different spatial scales $\mathcal{R}(B_T,\mathcal{E}_{\mathcal{T}})=1$ (LR) and $\mathcal{R}(B_T,\mathcal{E}_{\mathcal{T}})=4$ (HR) on the Ev-REDS dataset.}
% \vspace{-1em}
% \begin{tabular}{c|cccc|c}
% \hline
% ID & $\mathcal{L}_{BC}$        & $\mathcal{L}_{SC}$      & $\mathcal{L}_{SG}$       & $\mathcal{L}_{TG}$        &    LR /  HR PSNR $\uparrow$           \\ \hline
% \#1  & \checkmark &           &           &           & 19.43 / 19.95  \\
% \#2  &           & \checkmark &           &           & 12.75 / 13.90   \\
% \#3  & \checkmark & \checkmark &           &           & 21.67  / \underline{23.48}  \\
% \#4  & \checkmark & \checkmark & \checkmark &           & \underline{23.39} / 23.43  \\
% \#5  & \checkmark & \checkmark & \checkmark & \checkmark & \textbf{24.08} / \textbf{24.07}  \\ \hline
% \end{tabular}
% \vspace{-1em}
% \label{tab:ablation}
% \end{table}

\par
Regarding the case with different temporal scales, previous approaches are often limited by the blur distribution of training data, resulting in a significant performance drop when encountering large motion blur, as 
shown in Tab.~\ref{tab:ev-hsergb-main} and Fig.~\ref{fig:hs-ergb}. 
% Even though our GEM is only trained over the dataset with normal blur, the temporal generalization technique 
Benefiting from the temporal generalization technique in our learning framework, the proposed method can recover reliable latent images of the target scenes under both normal and large blur as depicted in Fig.~\ref{fig:hs-ergb}. 
Thus, our method not only enables flexible setups of input spatial resolution but also exhibits promising performance in handling motion blur of different temporal scales, facilitating applications in real-world scenarios.

% Regarding the case with different temporal scales, previous approaches are often confined to the blur distribution of their training data, resulting in a performance drop when encountering large motion blur, as 
% shown in Tab.~\ref{tab:ev-hsergb-main} and Fig.~\ref{fig:hs-ergb}. 
% Benefiting from the temporal generalization technique in our learning framework, the proposed method can recover reliable latent images of the target scenes under both normal and large blur as depicted in Fig.~\ref{fig:hs-ergb}. 
% Thus, our method not only enables flexible setups of input spatial resolution but also exhibits promising performance when handling motion blur of different temporal scales, facilitating applications in real-world scenarios.

\subsection{\textcolor{black}{Ablation Study}}
We study the contribution of each component in our self-supervised learning method on the Ev-REDS dataset and draw the following conclusions: 
\par
\begin{figure}[t]
	\centering
	\begin{subfigure}{\linewidth}
    \centering
    \begin{subfigure}{0.3865\linewidth}
    \begin{overpic}[width=\linewidth]
    {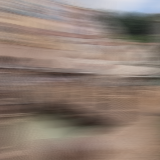}
    \put(1,4){\footnotesize \textcolor{white}{\bf LR blur}}
    \end{overpic}
    \end{subfigure}
    \begin{subfigure}{0.192\linewidth}
    \begin{overpic}[width=\linewidth]{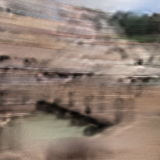}
        \put(1,4){\footnotesize \textcolor{white}{\bf \#1}}
        \end{overpic}
    \\
    \begin{overpic}[width=\linewidth]{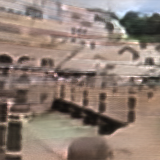}
        \put(1,4){\footnotesize \textcolor{white}{\bf \#4}}
        \end{overpic}
    \end{subfigure}
    \begin{subfigure}{0.192\linewidth}
    \begin{overpic}[width=\linewidth]{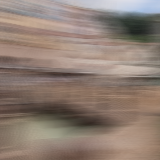}
        \put(1,4){\footnotesize \textcolor{white}{\bf \#2}}
        \end{overpic}
    \\
    \begin{overpic}[width=\linewidth]{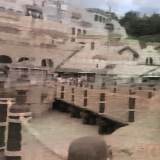}
        \put(1,4){\footnotesize \textcolor{white}{\bf \#5}}
        \end{overpic}
    \end{subfigure}
    \begin{subfigure}{0.192\linewidth}
    \begin{overpic}[width=\linewidth]{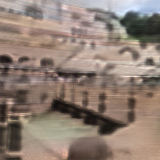}
        \put(1,4){\footnotesize \textcolor{white}{\bf \#3}}
        \end{overpic}
    \\
    \begin{overpic}[width=\linewidth]{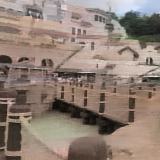}
        \put(1,4){\footnotesize \textcolor{white}{\bf \#6}}
        \end{overpic}
    \end{subfigure}
    \\
    %%% -----------------------------
    \begin{subfigure}{0.3865\linewidth}
    \begin{overpic}[width=\linewidth]
    {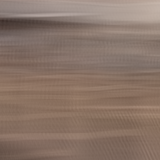}
    \put(1,4){\footnotesize \textcolor{white}{\bf HR blur}}
    \end{overpic}
    \end{subfigure}
    \begin{subfigure}{0.192\linewidth}
    \begin{overpic}[width=\linewidth]{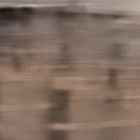}
        \put(1,4){\footnotesize \textcolor{white}{\bf \#1}}
        \end{overpic}
    \\
    \begin{overpic}[width=\linewidth]{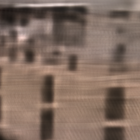}
        \put(1,4){\footnotesize \textcolor{white}{\bf \#4}}
        \end{overpic}
    \end{subfigure}
    \begin{subfigure}{0.192\linewidth}
    \begin{overpic}[width=\linewidth]{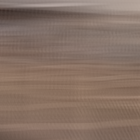}
        \put(1,4){\footnotesize \textcolor{white}{\bf \#2}}
        \end{overpic}
    \\
    \begin{overpic}[width=\linewidth]{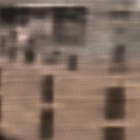}
        \put(1,4){\footnotesize \textcolor{white}{\bf \#5}}
        \end{overpic}
    \end{subfigure}
    \begin{subfigure}{0.192\linewidth}
    \begin{overpic}[width=\linewidth]{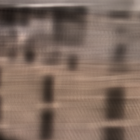}
        \put(1,4){\footnotesize \textcolor{white}{\bf \#3}}
        \end{overpic}
    \\
    \begin{overpic}[width=\linewidth]{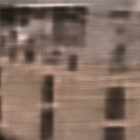}
        \put(1,4){\footnotesize \textcolor{white}{\bf \#6}}
        \end{overpic}
    \end{subfigure}
	% \begin{overpic}[width=0.394\linewidth]
 %    {imgs/main/Ablation/Blur-4x.png}
 %    \put(1,4){\footnotesize \textcolor{white}{\bf HR blur}}
 %    \end{overpic}
 %    \begin{subfigure}{\linewidth}
 %        \centering
 %        \begin{overpic}[width=0.195\linewidth]{imgs/main/Ablation/model1-4x.png}
 %        \put(1,4){\footnotesize \textcolor{white}{\bf \#1}}
 %        \end{overpic}
 %        \begin{overpic}[width=0.195\linewidth]{imgs/main/Ablation/model2-4x.png}
 %        \put(1,4){\footnotesize \textcolor{white}{\bf \#2}}
 %        \end{overpic}
 %        \begin{overpic}[width=0.195\linewidth]{imgs/main/Ablation/model3-4x.png}
 %        \put(1,4){\footnotesize \textcolor{white}{\bf \#3}}
 %        \end{overpic}
 %        \\
 %        \begin{overpic}[width=0.195\linewidth]{imgs/main/Ablation/model4-4x.png}
 %        \put(1,4){\footnotesize \textcolor{white}{\bf \#4}}
 %        \end{overpic}
 %        \begin{overpic}[width=0.195\linewidth]{imgs/main/Ablation/model5-4x.png}
 %        \put(1,4){\footnotesize \textcolor{white}{\bf \#5}}
 %        \end{overpic}
 %        \begin{overpic}[width=0.195\linewidth]{imgs/main/Ablation/model6-4x.png}
 %        \put(1,4){\footnotesize \textcolor{white}{\bf \#6}}
 %        \end{overpic}
 %    \end{subfigure}
	\caption{\rmfamily \fontsize{8pt}{0} Qualitative comparisons}
	\label{fig:ablation-quali-res}
	\end{subfigure}
	\begin{subfigure}{0.48\linewidth}
	    \includegraphics[width=\linewidth]{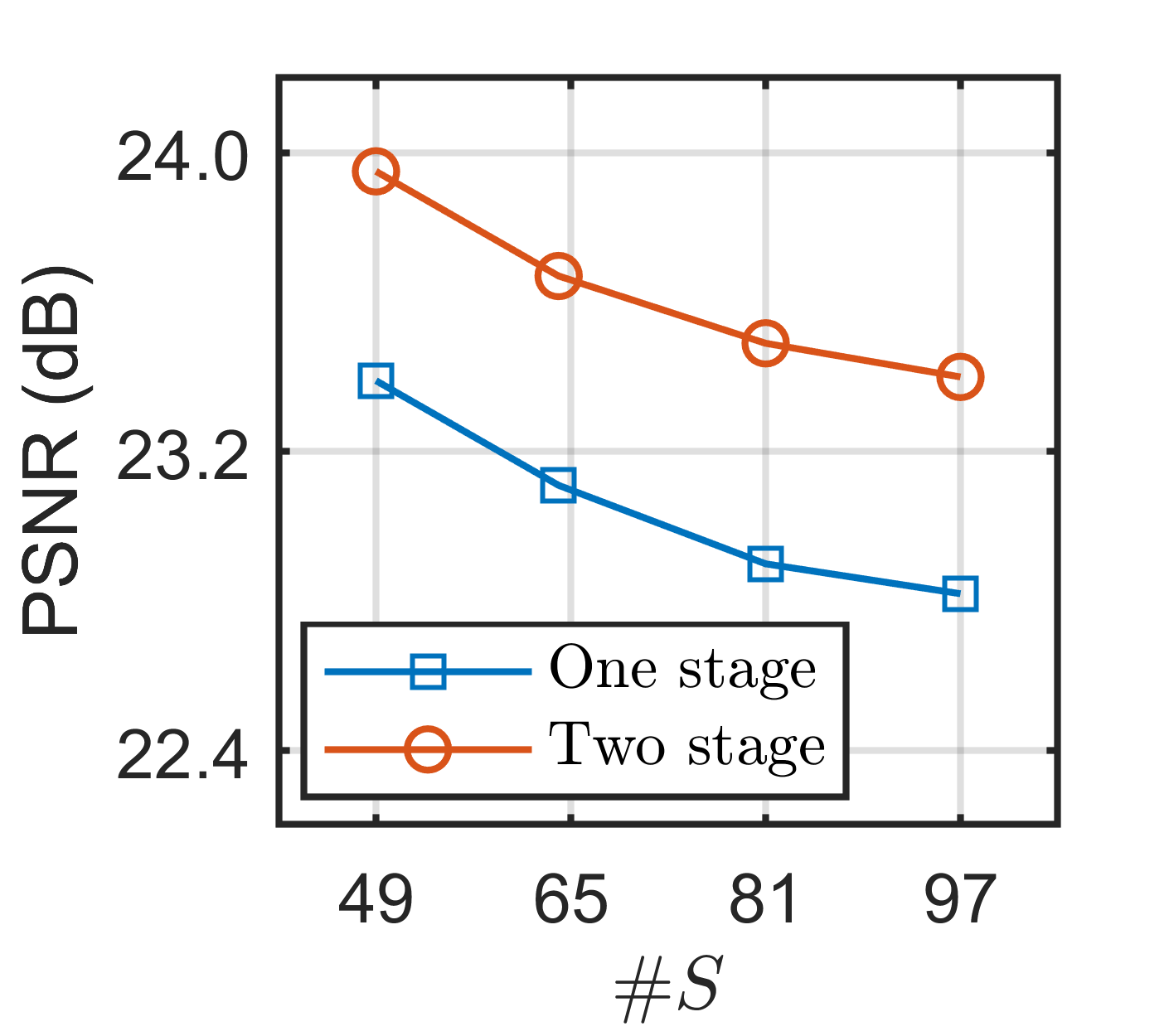}
	    \caption{\rmfamily \fontsize{8pt}{0} Temporal comparisons}
	    \label{fig:ablation-temporal-curve}
	\end{subfigure}
	\begin{subfigure}{0.48\linewidth}
	    \includegraphics[width=\linewidth]{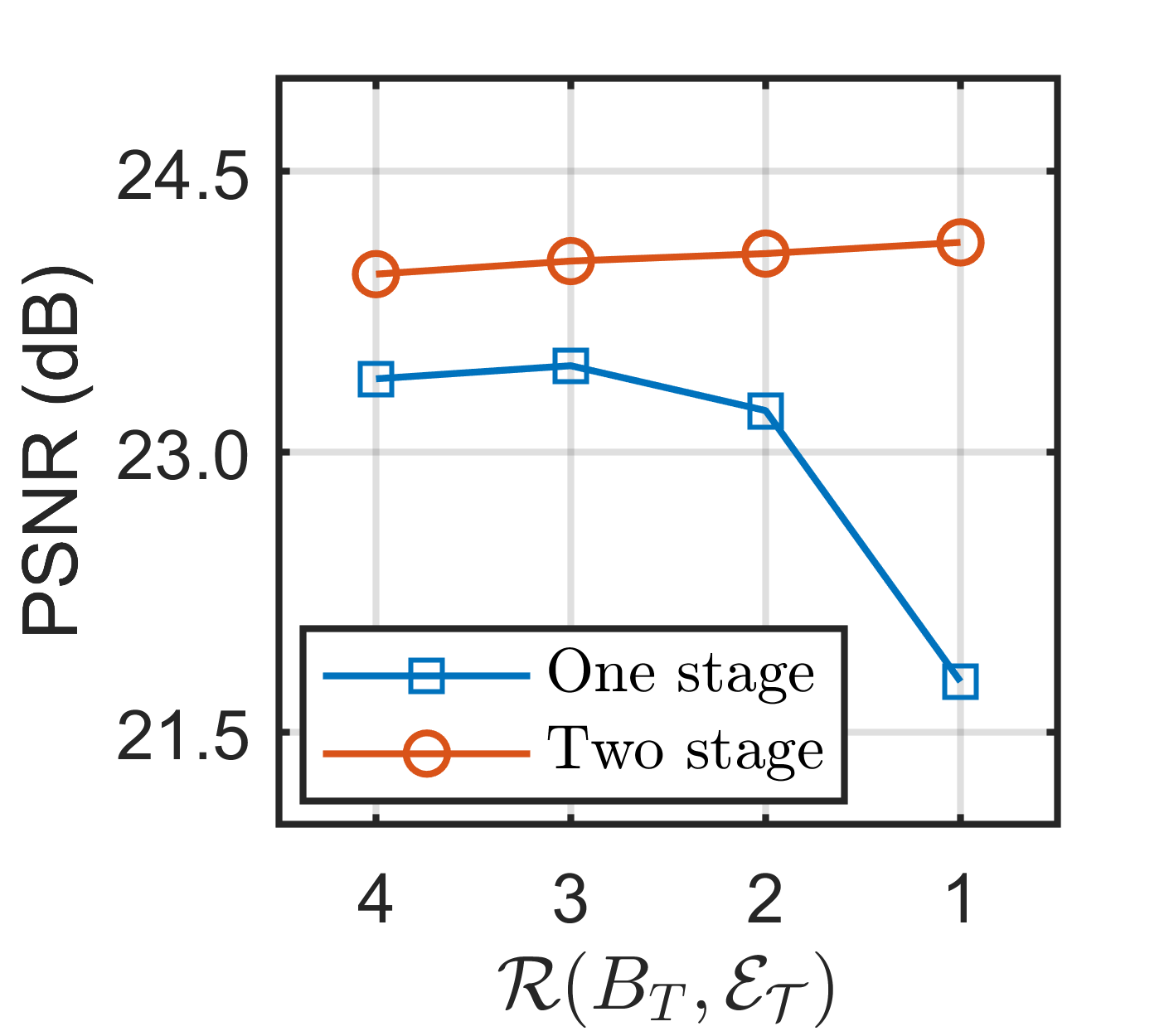}
	    \caption{\rmfamily \fontsize{8pt}{0} Spatial comparisons}
	    \label{fig:ablation-spatial-curve}
	\end{subfigure}
	% \vspace{-.75em}
	\caption{Results of different models in Tab.~\ref{tab:ablation} on the Ev-REDS dataset. (a) Qualitative comparisons under $\mathcal{R}(B_T,\mathcal{E}_{\mathcal{T}})=1$ (LR blur, top row) and $\mathcal{R}(B_T,\mathcal{E}_{\mathcal{T}})=4$ (HR blur, bottom row). (b, c) Comparisons of models using one-stage and two-stage training, \ie, model \#3 and \#6, under different temporal and spatial scales of motion blur.  $\#S$ denotes the number of sharp images used to synthesize one blurry frame, and larger $\#S$ indicates more blurred frames.
	}
	\label{fig:ablation}
        \vspace{-1em}
\end{figure}

 
\textbf{Combination of Brightness and Structure Consistency.} 
In Tab.~\ref{tab:ablation} and Fig.~\ref{fig:ablation-quali-res}, model \#1 trained only with $\mathcal{L}_{BC}$ effectively constrains the restored brightness by learning blur2blur conversion, but it suffers from missing structure and produces blurry results. By combining $\mathcal{L}_{BC}$ and $\mathcal{L}_{SC}$, model \#3 successfully recovers the correct structure with accurate brightness as shown in Fig.~\ref{fig:ablation-quali-res}, simultaneously guaranteeing brightness and structure consistency. 
However, training solely with $\mathcal{L}_{SC}$ will lead to  collapsing solutions as the results of model \#2 depicted in Fig.~\ref{fig:ablation-quali-res}. This is because the supervision signal in $\mathcal{L}_{SC}$ is strongly dependent on $\mathcal{L}_{BC}$ as discussed in Sec.~\ref{sec:ssl}, and thus the structure constraint $\mathcal{L}_{SC}$ should be used together with the brightness constraint $\mathcal{L}_{BC}$ to achieve motion deblurring. 

% Note that the structure constraint $\mathcal{L}_{SC}$ should be used together with the brightness constraint $\mathcal{L}_{BC}$ since the supervision signal in $\mathcal{L}_{SC}$ is strongly dependent on $\mathcal{L}_{BC}$ as discussed in Sec.~\ref{sec:ssl}, and training solely with $\mathcal{L}_{SC}$ will lead to trivial solutions as the results of model \#2 depicted in Fig.~\ref{fig:ablation-quali-res}.

\par 
\textbf{Effectiveness of Temporal and Spatial Generalization.} 
Although the first stage of training achieves promising performance in deblurring HR blurry frames with LR events, \ie, $\mathcal{R}(B_T,\mathcal{E}_{\mathcal{T}})=4$, it struggles to handle different temporal and spatial scales of motion blur as shown in Fig.~\ref{fig:ablation-temporal-curve} and \ref{fig:ablation-spatial-curve}. To improve the deblurring performance in the temporal dimension,  our $\mathcal{L}_{TG}$ supervises the consistency of latent images restored from blurry frames with different levels of blurriness.
Since it is generally easier to deblur the frames with normal blur ($B_T$) than that with large blur ($B_{\tilde{T}}$), $\mathcal{L}_{TG}$ encourages our model to produce similar results when dealing with both cases and thus learns to tackle more severe motion blur, leading to better deblurring performance (models \#3 and \#4 in Tab.~\ref{tab:ablation} and Fig.~\ref{fig:ablation-quali-res}) and general improvements in large blur removal (Fig.~\ref{fig:ablation-temporal-curve}). 
In the spatial domain, the performance inconsistency shown in Tab.~\ref{tab:ablation} and Fig.~\ref{fig:ablation-spatial-curve} is because model \#3 only learns to project events to fit HR frames but neglects the varying event distributions at different spatial scales.  With $\mathcal{L}_{SG}$, our SAN can adaptively adjust the learned event distribution according to $\mathcal{R}(B_T,\mathcal{E}_{\mathcal{T}})$ and propagate the deblurring performance under $\mathcal{R}(B_T,\mathcal{E}_{\mathcal{T}})=4$ to other spatial scales, leading to consistent performance as shown in Fig.~\ref{fig:ablation-quali-res} and \ref{fig:ablation-spatial-curve}.

\def\scc{(-0.65,-0.5)}
\def\imgWidth{0.158\linewidth} %子图大小

\section{Conclusion}
This paper proposes to generalize event-based motion deblurring in real-world scenarios. We first present a scale-aware network to allow flexible setups of input spatial resolutions and enable learning from different temporal scales of motion blur. Following that, a two-stage self-supervised learning framework is designed for model training with real data and performance generalization in both spatial and temporal domains.
In addition, a real-world dataset containing high-resolution blurry frames and low-resolution events is released to facilitate the evaluation of frame/event-based deblurring approaches in real-world scenes.

{\small
\bibliographystyle{ieee_fullname}
\bibliography{egbib}
}

\end{document}